\ifdefined\pdfoutput
  \pdfoutput=1
\fi
\documentclass{article}

\usepackage{iftex}
\ifPDFTeX
\else\ifXeTeX
  
\fi\fi
\usepackage[OT1]{fontenc}

\usepackage[table]{xcolor}

\PassOptionsToPackage{numbers, compress}{natbib}
\usepackage[main, final]{neurips_2026}

\definecolor{darkblue}{rgb}{0, 0, 0.5}
\definecolor{forestgreen}{RGB}{34,139,34}

\usepackage{microtype}
\usepackage{graphicx}
\usepackage{subcaption}
\usepackage{booktabs}
\usepackage{tabularx}
\usepackage{multirow}
\usepackage{enumitem}

\usepackage{hyperref}
\hypersetup{colorlinks=true, citecolor=darkblue, linkcolor=darkblue, urlcolor=darkblue}

\usepackage{amsmath}
\usepackage{amssymb}
\usepackage{mathtools}
\usepackage{amsthm}

\usepackage[capitalize,noabbrev]{cleveref}

\usepackage{titletoc}
\newcommand\DoToC{%
  \startcontents
  \begingroup
  \setcounter{tocdepth}{2}
  \printcontents{}{1}{\textbf{Contents of Appendix}\vskip3pt\hrule\vskip5pt}
  \vskip3pt\hrule\vskip5pt
  \endgroup
}

\usepackage{algorithm}
\usepackage{algorithmic}

\usepackage{fontawesome5}

\usepackage{pifont}

\theoremstyle{plain}

\theoremstyle{definition}

\theoremstyle{remark}

\title{MemReward: Graph-Based Experience Memory for LLM Reward Prediction with Limited Labels}

\author{%
  Tianyang Luo$^{*1}$ \quad Tao Feng$^{*1}$ \quad Zhigang Hua$^{2}$ \quad Yan Xie$^{2}$\\
  \textbf{Shuang Yang}$^{2}$ \quad \textbf{Ge Liu}$^{1}$ \quad \textbf{Jiaxuan You}$^{1\dagger}$\\[0.3em]
  $^{1}$University of Illinois Urbana-Champaign \qquad $^{2}$Meta\\[0.2em]
  {\small $^{*}$Equal contribution. $^{\dagger}$Correspondence to: \texttt{jiaxuan@illinois.edu}}
}

\begin{document}

\maketitle

\vspace{-0.2in}
\begin{center}
{\large \faGithub~\href{https://github.com/ulab-uiuc/MemReward}{ulab-uiuc/MemReward}\qquad \includegraphics[height=1em]{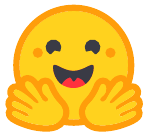}~\href{https://huggingface.co/datasets/ulab-ai/memreward}{ulab-ai/memreward}}
\end{center}
\vspace{-0.1in}

\begin{abstract}
Reinforcement learning (RL) fine-tuning improves large language model (LLM) reasoning using reward signals on sampled rollouts. When reward labels are available for only a subset of rollouts, restricting policy updates to that subset leaves the remaining rollouts unused for policy optimization. Semi-supervised learning propagates labels from labeled to unlabeled samples, but existing methods either diffuse labels with fixed similarity weights or treat each sample independently, ignoring the relations among queries, thinking processes, and answers. We propose \textbf{MemReward}, a graph-based experience memory that propagates rewards to unlabeled rollouts during online policy optimization. MemReward stores queries and the rollouts (thinking processes and final answers) of an initial LLM policy as nodes of a heterogeneous graph connected by similarity and structural edges, and warms up a GNN on labeled rollouts to propagate rewards over this graph. During online RL fine-tuning, unlabeled rollouts are attached to the graph by query similarity, and the policy is updated with ground-truth rewards on labeled rollouts and GNN-predicted rewards on the rest. We evaluate MemReward on 13 benchmarks in mathematics, question answering, and code generation with verifiable rewards, and further with LLM-judge, human-preference, and continuous rewards. With ground-truth rewards on only 20\% of rollouts, MemReward achieves 96.6\% of the performance of fully supervised training (Oracle) on Qwen2.5-1.5B and 97.3\% on Qwen2.5-3B, where it also outperforms label propagation and FixMatch-style self-training. It carries over to Llama backbones up to 8B and closely approaches Oracle on out-of-domain tasks.
\end{abstract}

\section{Introduction}
\label{sec:intro}
Recent advances in large language models (LLMs) have been driven by reinforcement-learning-based post-training, where rewards on sampled rollouts guide policy optimization~\citep{guo2025deepseek,yu2025dapo,cui2025process,jin2025search}. We study a limited-supervision setting in which reward labels are available for only a subset of sampled rollouts. Training solely on this labeled subset discards potentially useful unlabeled rollouts. Our goal is to exploit these unlabeled rollouts while keeping the reward-label budget fixed. A natural solution lies in semi-supervised learning, where models trained on labeled data have proven effective in propagating labels to unlabeled samples~\citep{lee2013pseudo,iscen2019label,sohn2020fixmatch}. This suggests an analogous strategy for RL fine-tuning: treating labeled rollouts and their rewards as early experiences, linking unlabeled rollouts to these experiences, and propagating rewards from labeled to unlabeled rollouts. Therefore, we ask: \textit{Can we leverage early experience memories to construct a shared reward model, enabling effective RL fine-tuning in label scarcity scenarios?}

\begin{figure}[t]
\centering
\includegraphics[width=1.0\textwidth]{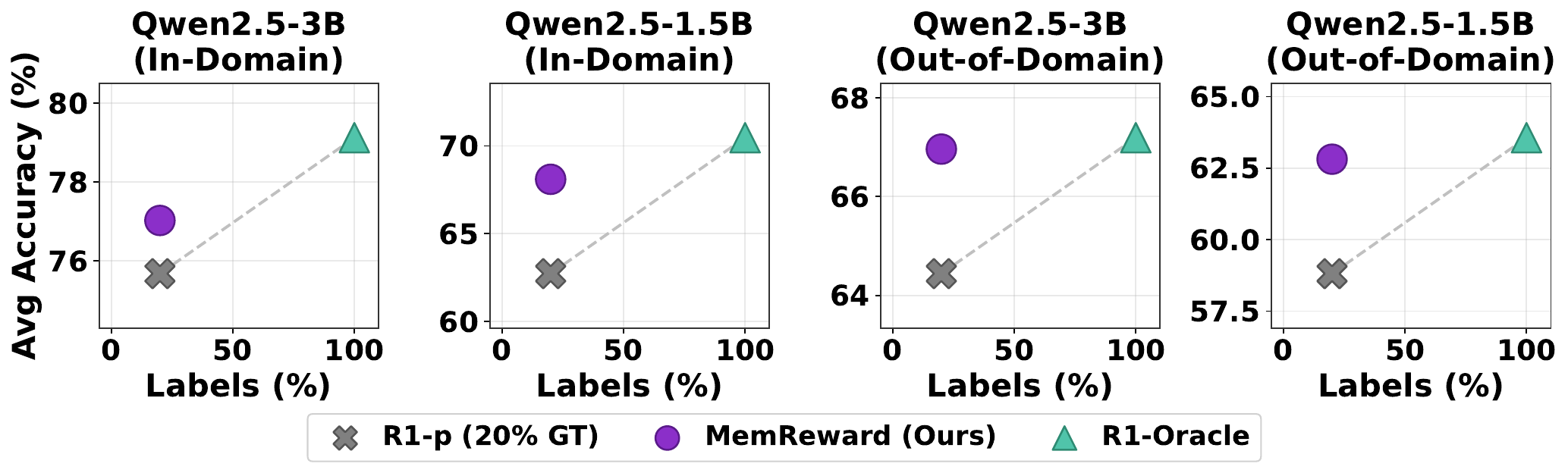}
\caption{\textbf{MemReward approaches Oracle performance with only 20\% labels.} Using the same 20\% ground-truth labels, MemReward (purple) substantially outperforms partial labels (R1-p, gray), closely approaching fully-supervised Oracle performance (green) on both in-domain and out-of-domain tasks across both model scales.}
\label{fig:teaser}
\vspace{-3mm}
\end{figure}

Addressing this question presents several challenges. First, the relationships among experiences are complex and multi-faceted: semantically similar queries may share reward patterns, multiple rollouts for the same query compete in quality, and answer correctness depends on both the reasoning path and the query context. Modeling such intertwined dependencies calls for a relational structure such as a graph, rather than treating each experience independently~\citep{qiu2025reward,huang2021combining}. Second, the relevant relations are themselves heterogeneous: spanning query-query semantic similarity, query-to-reasoning association, and reasoning-to-answer correspondence---and a homogeneous graph cannot distinguish their differing semantics, motivating a heterogeneous design~\citep{wang2019heterogeneous,schlichtkrull2018modeling}.

To address these challenges, we introduce \textbf{MemReward}, a graph-based experience memory framework. We first use an initial LLM policy to generate rollouts for each labeled query, where each rollout consists of a thinking process and a final answer. Queries, thinking processes, and answers form three types of nodes, connected by three types of edges: query-query edges between semantically similar queries, query-thinking edges from each query to its rollouts, and thinking-answer edges within each rollout. A heterogeneous GNN is then trained jointly across multiple domains (mathematics, QA, and code generation) over this graph to predict rewards~\citep{gilmer2017neural,kipf2016semi}. During online RL fine-tuning, unlabeled rollouts are attached to the graph via embedding similarity, and the GNN predicts their rewards to augment the limited ground-truth signal during policy optimization.

We evaluate MemReward on Qwen2.5-1.5B and 3B across 13 benchmarks spanning 3 domains: mathematics, question answering, and code generation. As shown in Figure~\ref{fig:teaser}, MemReward---using ground-truth rewards on only 20\% of rollouts and GNN-predicted rewards on the rest---achieves 96.6\% of Oracle performance on 1.5B and 97.3\% on 3B. On out-of-domain tasks, MemReward closely approaches Oracle on both model scales, showing that GNN-predicted rewards generalize effectively across domain boundaries. Performance scales smoothly with the label budget, reaching 99.4\% of Oracle at 70\% labels. Beyond these settings, MemReward improves over R1-p with Llama-3.2-3B and Llama-3.1-8B backbones, with LLM-judge (TruthfulQA), human-preference (Chatbot Arena), and continuous token-F1 (HotpotQA) rewards, and when the GNN is trained on one domain and used for RL in another. Across six random seeds, its gains over R1-p on both Qwen model sizes are statistically significant. Our conclusions concern label efficiency in the evaluated reward settings, while expert-graded mathematical proofs remain a direction for future work (Appendix~\ref{appendix:limitations}).

\section{Preliminaries}
\label{sec:preliminaries}

\textbf{Reinforcement Learning for Reasoning-based LLM Fine-tuning.}
In recent RL-based fine-tuning, an LLM policy $\pi_\theta$ parameterized by $\theta$ generates rollouts $y$ for a query $q$~\citep{guo2025deepseek}. Each rollout (also referred to as an \textit{experience}) consists of a thinking process $t$ and a final answer $a$. The correctness of the final answer $a$ determines the reward, and the policy is updated through policy gradient methods to reinforce the entire thinking process $t$ that produced a correct answer.

\textbf{Group Relative Policy Optimization (GRPO).}
GRPO~\citep{guo2025deepseek} is a widely adopted RL algorithm for reasoning-based LLM fine-tuning. For each query, it compares the results of multiple rollouts to obtain the advantage values. It can be formulated as follows:

\vspace{-0.3em}
\begin{equation}
\mathcal{J}_{\text{GRPO}}(\pi_\theta) = \mathbb{E} \Big[\frac{1}{N}\sum_{i=1}^{N}\frac{1}{|y_i|}\sum_{j=1}^{|y_i|}\min\{\rho_{i,j}A_i, \text{clip}(\rho_{i,j}, 1-\epsilon, 1+\epsilon)A_i\} - \beta \text{KL}(\pi_\theta\|\pi_{\text{ref}})\Big]
\label{eq:grpo}
\end{equation}
\vspace{-0.3em}

where $q \sim \mathcal{D}$ is a query sampled from the data distribution; $y_{1...N} \sim \pi_\theta(q)$ are $N$ rollouts generated by the policy; $|y_i|$ is the token length of the rollout $y_i$; $\rho_{i,j}=\pi_\theta(y_{i,j}|q,y_{i,<j})/\pi_{\text{old}}(y_{i,j}|q,y_{i,<j})$ is the importance sampling ratio for the $j$-th token of rollout $i$; $A_i = \frac{R(q,y_i) - \text{mean}[R(q,y_{1...N})]}{\text{std}[R(q,y_{1...N})]}$ is the standardized advantage; $R(q, y_i)$ is the reward for the rollout $y_i$; $\epsilon$ is the clipping range; $\beta$ is the KL penalty weight; and $\pi_{\text{ref}}$ is the reference policy.

\section{MemReward: Graph-Based Experience Memory for Reward Prediction}

\begin{figure}[t]
    \centering
    \includegraphics[width=1.0\textwidth]{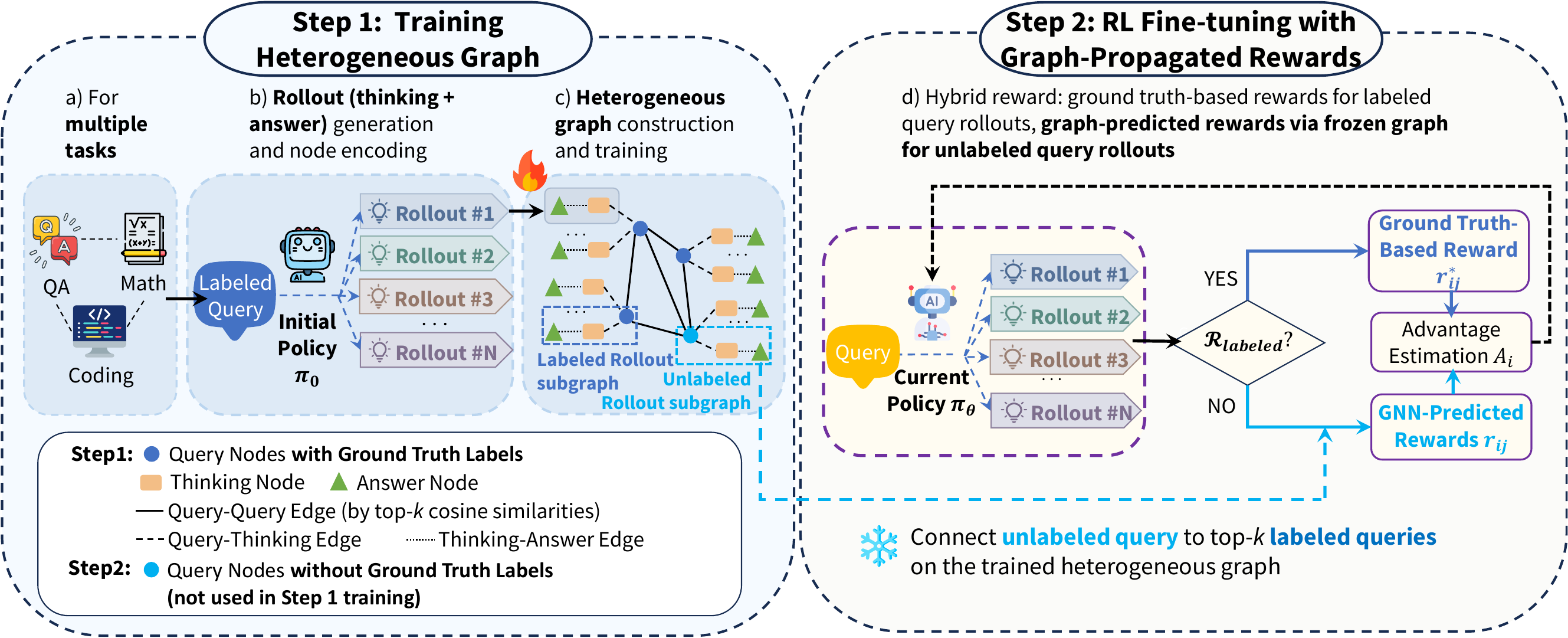}
    \caption{\textbf{Overview of MemReward.} Rollouts generated by the initial policy are stored as experience memory and organized into a heterogeneous graph for reward prediction. (Left) \textit{Warmup Phase}: We construct a heterogeneous graph from labeled rollouts, where query nodes connect via embedding similarity, and each query links to its thinking and answer nodes. A GNN is trained to predict rewards through relational message passing. (Right) \textit{Online Phase}: During GRPO training, labeled rollouts receive ground-truth rewards while unlabeled rollouts connect to the warmup graph via top-$k$ similarity edges and obtain GNN-predicted rewards.}
    \label{fig:architecture}
    \vspace{-3mm}
\end{figure}

\textbf{Overview.} As illustrated in Figure~\ref{fig:architecture}, MemReward treats each rollout---consisting of a query, a thinking process, and an answer---as a labeled experience stored in a graph-based memory. It constructs a cross-domain heterogeneous graph from these labeled experiences and trains a GNN to predict rewards (Section~\ref{sec:cross_domain_graph}). During online policy optimization, the trained GNN predicts rewards for unlabeled rollouts, which are combined with ground-truth rewards from labeled rollouts for scalable reward acquisition (Algorithm~\ref{alg:grpo_online}).

\subsection{Cross-Domain Heterogeneous Graph for Reward Prediction}
\label{sec:cross_domain_graph}
\textbf{Initialize node/edge features.} In MemReward, we have three types of nodes (query nodes, thinking nodes, and answer nodes) and three types of edges (query-query, query-thinking, and thinking-answer). Their embeddings at layer $\ell$ are denoted as $h_q^{(\ell)}$, $h_t^{(\ell)}$, and $h_a^{(\ell)}$, with direction-specific weight matrices $W_{qq}$, $W_{tq}$, $W_{qt}$, $W_{at}$, and $W_{ta}$.

For node initialization, for each query $q$ with ground-truth labels in the training set, we use the initial policy $\pi_0$ to generate rollouts, each consisting of a thinking process $t$ and a final answer $a$, and then encode the query, thinking, and answer to obtain embeddings $e_q, e_t, e_a$.

For edge construction, query-query edges are established using the top-$k$ cosine similarity between query embeddings. Query-thinking edges directly connect each query node to its thinking nodes. Thinking-answer edges pair each thinking node one-to-one with the corresponding answer node. All edge weights are initialized to 1. Only query-query edges depend on embedding similarity; thinking and answer nodes enter the graph solely through structural edges within their own rollout, so their length or template does not affect which experiences are retrieved.

The graph represents relations among experiences rather than temporal dependencies. A directed acyclic graph (DAG) variant could orient query-query edges by generation time or label confidence, but query similarity does not define a global acyclic ordering. Enforcing one would introduce an ordering heuristic and could remove reciprocal message-passing paths between related experiences. We therefore retain the heterogeneous graph without imposing acyclicity and leave DAG-based variants for future work.

\textbf{Training the warmup GNN.} We train a heterogeneous GNN on labeled rollouts to enable cross-query knowledge transfer, as shown in Figure~\ref{fig:architecture}.

For aggregating different types of nodes and edges, we employ heterogeneous aggregation with type-specific weights. The GNN derives expressive node representations $h$ through iterative neighborhood aggregation with learnable weights. The update of the embedding of the node in the $\ell$-th layer is as follows:

\vspace{-0.5em}
{\footnotesize
\begin{equation}
\mathbf{h}_q^{(\ell)} = \text{ReLU}\Big(\text{Mean}\big(\sum_{\mathcal{N}_q} \alpha^{qq} W_{qq}^{(\ell)} \mathbf{h}_q^{(\ell-1)}, \sum_{\mathcal{N}_t} \alpha^{tq} W_{tq}^{(\ell)} \mathbf{h}_t^{(\ell-1)}\big)\Big),
\label{eq:h_q}
\end{equation}}
\vspace{-0.5em}
{\footnotesize
\begin{equation}
\mathbf{h}_t^{(\ell)} = \text{ReLU}\Big(\text{Mean}\big(\sum_{\mathcal{N}_q} \alpha^{qt} W_{qt}^{(\ell)} \mathbf{h}_q^{(\ell-1)}, \sum_{\mathcal{N}_a} \alpha^{at} W_{at}^{(\ell)} \mathbf{h}_a^{(\ell-1)}\big)\Big),
\label{eq:h_t}
\end{equation}}
\vspace{-0.5em}
{\footnotesize
\begin{equation}
\mathbf{h}_a^{(\ell)} = \text{ReLU}\Big(\sum_{\mathcal{N}_t} \alpha^{ta} W_{ta}^{(\ell)} \mathbf{h}_t^{(\ell-1)}\Big),
\label{eq:h_a}
\end{equation}}
\vspace{-0.5em}

where $\mathbf{h}^{(\ell)}$ is the embedding of the node after $\ell$ iterations.
$\mathbf{h}_q^{(\ell)}$, $\mathbf{h}_t^{(\ell)}$,
$\mathbf{h}_a^{(\ell)}$ are initialized as
$\mathbf{h}_q^{(0)} = e_q$, $\mathbf{h}_t^{(0)} = e_t$,
$\mathbf{h}_a^{(0)} = e_a$.
$\alpha^{xy}$ denotes the attention weights for edge type $x \to y$, $W_{xy}^{(\ell)}$ denotes the learnable edge type-specific weights, and $\mathcal{N}_x$ denotes the neighbors of node type $x$.

Based on the final-layer embeddings for the query, thinking, and answer nodes, we predict the reward score for each rollout using scaled dot product scoring:

\begin{equation}
r_{ij} = \frac{\phi_q\big(h_{q_{i}}^{(L)}\big) \cdot \phi_r\big([h_{t_{ij}}^{(L)} \| h_{a_{ij}}^{(L)}]\big)}{\sqrt{d}} + b,
\label{eq:gnn_reward}
\end{equation}

where $r_{ij}$ is the predicted reward score for the $j$-th rollout of query $i$; $\phi_q$ and $\phi_r$ are learnable linear projections that map the query and rollout representations to a shared $d$-dimensional space, respectively; $h_{q_i}^{(L)}$ is the final-layer embedding of the $i$-th query; $h_{t_{ij}}^{(L)}$ and $h_{a_{ij}}^{(L)}$ are the final-layer embeddings of the thinking process and answer for the $j$-th rollout of query $i$; $\|$ denotes concatenation; and $b$ is a learnable bias.


The GNN is trained to minimize the binary cross-entropy loss:

\begin{equation}
\mathcal{L}_{\text{BCE}} = -\frac{1}{N} \sum_{i,j} \big[ r_{ij}^{*} \log \sigma(r_{ij}) + (1-r_{ij}^{*}) \log(1-\sigma(r_{ij}))\big],
\label{eq:loss_bce}
\end{equation}
\vspace{-0.5em}

\noindent where $\sigma(\cdot)$ is the sigmoid function; $r_{ij}^{*} = r(a_{ij}, a_{i}^*) \in \{0,1\}$ is the ground-truth reward, equal to 1 if the answer is correct and 0 otherwise; $r(\cdot, \cdot)$ is the reward function that compares an answer with the ground-truth label; $a_{ij}$ is the $j$-th generated answer for the $i$-th query; $a_{i}^*$ is its ground-truth label; and $N$ is the total number of rollouts for each query. For continuous rewards such as token-level F1, we instead train the GNN with a regression loss to predict the reward directly.

\subsection{Adaptive Reward Acquisition for Online Policy Optimization}

\textbf{Using the trained GNN.} During online policy optimization,
we employ a mixed reward acquisition strategy: labeled rollouts receive ground-truth rewards, while unlabeled rollouts leverage GNN-predicted rewards.

Since unlabeled rollouts and their associated queries are not part of the warmup graph, we connect them to the warmup graph (Section~\ref{sec:cross_domain_graph}) at inference time. For each query $q$ whose rollouts are unlabeled, we first encode it and its rollouts using the same encoder to obtain embeddings $e_{q}$, $e_{t}$, and $e_{a}$. We then establish query-query edges between the new query embedding and its top-$k$ most similar warmup query embeddings by cosine similarity. The initial embeddings are then propagated through the trained GNN to obtain the final layer representations $h_{q}^{(L)}$, $h_{t}^{(L)}$, and $h_{a}^{(L)}$. Finally, we predict the reward of rollout $j$ for query $i$ as shown in Equation~\eqref{eq:gnn_reward}.

\textbf{Integrating GNN with GRPO.}
During online policy optimization, for each query $q_i$, GRPO generates $N$ rollouts $\{y_1, \ldots, y_N\}$ from the policy. The final reward for each rollout is computed as follows:

\begin{equation}
R_{ij} = \begin{cases}
r_{ij}^{*} & \text{if } y_{ij} \in \mathcal{R}_{\text{labeled}}, \\
\mathbb{I}[\sigma(r_{ij}) > 0.5] & \text{otherwise},
\end{cases}
\end{equation}

where $R_{ij}$ is the reward for the $j$-th rollout of query $i$, serving as $R(\cdot)$ in Equation~\eqref{eq:grpo} for computing the standardized advantage; $r_{ij}^{*}$ is the ground-truth reward and $r_{ij}$ is the GNN-predicted reward score; $\mathcal{R}_{\text{labeled}}$ denotes the set of labeled rollouts (i.e., rollouts whose queries have ground-truth labels); and $\mathbb{I}[\cdot]$ is the indicator function. For continuous rewards, the predicted score is used directly in place of the indicator.

The policy is then optimized by maximizing the GRPO objective in Equation~\eqref{eq:grpo}.

\begin{algorithm}[t]
\caption{Online GRPO with MemReward}
\label{alg:grpo_online}
\begin{algorithmic}[1]
\small
\STATE \textbf{Input:} $\mathcal{R}_{\text{labeled}}$ (labeled rollouts), $\mathcal{R}_{\text{unlabeled}}$ (unlabeled rollouts), trained GNN, $\mathcal{G}_{\text{warmup}}$, LLM policy $\pi_\theta$
\STATE \textbf{Output:} Fine-tuned policy $\pi_{\theta^*}$
\STATE \textbf{Prompt:} Domain-specific system prompt (see Appendix~\ref{appendix:case_studies})
\STATE
\STATE $\mathcal{R}_{\text{train}} \leftarrow \mathcal{R}_{\text{labeled}} \cup \mathcal{R}_{\text{unlabeled}}$
\FOR{each training iteration}
    \STATE Sample batch $\mathcal{B} \sim \mathcal{R}_{\text{train}}$
    \FOR{each query $q \in \mathcal{B}$}
        \STATE Generate $N$ rollouts: $\{(t_j, a_j)\}_{j=1}^N \sim \pi_\theta(\text{Prompt}, q)$
        \IF{$(t_j, a_j) \in \mathcal{R}_{\text{labeled}}$}
            \STATE $r_j \leftarrow r(a_j, a^*)$ \quad \textcolor{gray}{// Labeled rollout: ground-truth reward}
        \ELSE
            \STATE Encode: $e_q, e_{t_j}, e_{a_j}$
            \STATE Connect to $\mathcal{G}_{\text{warmup}}$ via top-$k$ similarity
            \STATE Propagate through GNN
            \STATE $r_j \leftarrow \mathbb{I}[\sigma(\text{GNN}(q, t_j, a_j)) > 0.5]$ \quad \textcolor{gray}{// Unlabeled rollout: GNN reward}
        \ENDIF
    \ENDFOR
    \STATE Compute advantages: $A_j = (r_j - \mu) / \text{std}$
    \STATE Update: $\theta \leftarrow \theta + \alpha \nabla_\theta \mathcal{J}_{\text{GRPO}}$
\ENDFOR
\STATE \textbf{Return} $\pi_{\theta^*}$
\end{algorithmic}
\end{algorithm}

\section{Experiments}

\subsection{Experimental Settings}
\textbf{Datasets.} We evaluate 13 datasets in total: 10 in-domain and 3 out-of-domain, spanning three task categories (math reasoning, question answering, and code generation).

\textbf{In-domain Benchmarks (10 datasets).} We evaluate in three task categories: \textbf{Math}: GSM8K~\citep{cobbe2021training}, GSM-Symbolic~\citep{mirzadeh2024gsm}, MATH~\citep{hendrycks2021measuring}; \textbf{Code}: MBPP+~\citep{austin2021program, liu2023your}, HumanEval+~\citep{chen2021evaluating, liu2023your}; \textbf{QA}: MMLU~\citep{hendrycks2020measuring}, CommonsenseQA~\citep{talmor2019commonsenseqa}, OpenBookQA~\citep{mihaylov2018can}, ARC-Challenge~\citep{clark2018think}, GPQA~\citep{rein2024gpqa}. Dataset statistics and detailed descriptions are given in the Appendices~\ref{appendix:dataset_statistics} and \ref{appendix:dataset_descriptions}.

\textbf{Out-of-domain Benchmarks (3 datasets).} To assess generalization, we evaluate three held-out datasets not seen during training: NuminaMath~\citep{li2024numinamath} (competition math), SIQA~\citep{sap2019socialiqa} (social reasoning) and PIQA~\citep{bisk2020piqa} (physical common sense).

\textbf{Baselines.}
All methods adopt GRPO~\citep{guo2025deepseek} as the RL algorithm for policy optimization, but differ in how rewards are acquired. We compare against six baselines: (1)~\textbf{R1-p (Partial)}, which trains with only 20\% labeled rollouts and discards the remaining 80\%; (2)~\textbf{R1-Oracle}, which trains with 100\% ground-truth labels, serving as a fully supervised reference; (3)~\textbf{kNN}, which predicts rewards for unlabeled rollouts by majority voting over the nearest labeled query embeddings; (4)~\textbf{LLM-RM}, which fine-tunes Qwen2.5-3B as a binary reward classifier via SFT on the 20\% labeled data and uses the frozen model to predict rewards for the remaining 80\%; (5)~\textbf{Label Propagation}~\citep{zhu2002learning}, which uses the same query embeddings, similarity graph, labeled rollouts, and GRPO procedure as MemReward but replaces learned heterogeneous message passing with fixed label diffusion; and (6)~\textbf{FixMatch-style self-training}~\citep{sohn2020fixmatch}, which assigns pseudo-rewards to the unlabeled rollouts on which the policy is most confident (Appendix~\ref{appendix:ssl}). In contrast, \textbf{MemReward} uses the same 20\% labeled rollouts as R1-p, but augments the remaining 80\% with GNN-predicted rewards.

\textbf{Settings.}
We use Qwen2.5-1.5B-Instruct and Qwen2.5-3B-Instruct as backbone models, and additionally Llama-3.2-3B-Instruct and Llama-3.1-8B-Instruct~\citep{grattafiori2024llama} (Section~\ref{sec:generalization}). For optimization of the online policy, we adopt GRPO with hyperparameters in Table~\ref{tab:hyperparams}. The GNN training configurations are shown in Table~\ref{tab:gnn_hyperparams} (complete implementation details are given in Appendix~\ref{appendix:hyperparams}). Case studies illustrating model output are provided in Appendix~\ref{appendix:case_studies}, and compute overhead is analyzed in Appendix~\ref{appendix:cost}.

\textbf{Evaluation Metrics.}
We use task-specific metrics across task categories: \textbf{Exact Match} for mathematical reasoning requiring precise numerical answers; \textbf{Accuracy} for question answering involving multiple-choice or classification; and \textbf{Pass@1} for code generation, measuring functional correctness.

\begin{table}[t]
\centering
\caption{\textbf{MemReward reaches 97.3\% of Oracle performance on Qwen2.5-3B and 96.6\% on Qwen2.5-1.5B across 10 benchmarks.} We compare reward acquisition strategies under a 20\% label budget. On 3B, MemReward outperforms R1-p by +1.35, LLM-RM by +1.30, and kNN and Label Propagation by +1.05 each. Label Propagation is evaluated on Qwen2.5-3B. \textbf{Bold} and \underline{underline} mark the best and second-best results among R1-p, MemReward, and R1-Oracle (a three-way tie is left unmarked). Avg is the mean over six random seeds, with the standard deviation as a subscript.}
\label{tab:main_results}

\small
\resizebox{1.0\textwidth}{!}{
\begin{tabular}{lcccccccccccc}
\toprule
\multirow{2}{*}{\textbf{Method}} & \multicolumn{3}{c}{\textbf{Math}} & \multicolumn{5}{c}{\textbf{QA}} & \multicolumn{2}{c}{\textbf{Code}} & \multirow{2}{*}{\textbf{Avg}} & \multirow{2}{*}{$\Delta \mathrm{Score}$} \\
\cmidrule(lr){2-4} \cmidrule(lr){5-9} \cmidrule(lr){10-11}
& GSM8K & GSM-sym & MATH & MMLU & CSQA & OBQA & ARC-C & GPQA & HumanEval+ & MBPP+ & & \\
\midrule
\multicolumn{13}{l}{\textit{Qwen2.5-1.5B-Instruct}} \\
\midrule[0.3pt]
R1-p & 77.11 & 62.89 & 44.44 & 53.33 & 70.22 & 68.67 & 71.56 & \underline{20.00} & 38.46 & \underline{55.00} & 62.72$_{\pm1.75}$ & -7.75 \\
LLM-RM & 78.22 & 64.00 & 45.35 & 54.44 & 70.34 & 68.79 & 72.22 & 18.33 & 41.03 & 55.00 & 63.29$_{\pm1.91}$ & -7.18 \\
kNN & 79.33 & 65.11 & 46.49 & 55.33 & 71.91 & 70.62 & 73.74 & 20.00 & 41.03 & 52.50 & 64.49$_{\pm1.58}$ & -5.98 \\
\rowcolor{violet!10} \textbf{MemReward (Ours)} & \textbf{88.67} & \textbf{77.78} & \underline{50.89} & \underline{54.67} & \underline{72.44} & \underline{70.00} & \underline{72.67} & \textbf{23.33} & \underline{43.59} & \underline{55.00} & \underline{68.10}$_{\pm1.62}$ & -2.37 \\
R1-Oracle & \underline{86.44} & \underline{75.33} & \textbf{53.11} & \textbf{66.44} & \textbf{74.44} & \textbf{74.00} & \textbf{74.89} & 15.00 & \textbf{53.85} & \textbf{56.25} & \textbf{70.47}$_{\pm1.54}$ & 0 \\
\midrule
\multicolumn{13}{l}{\textit{Qwen2.5-3B-Instruct}} \\
\midrule[0.3pt]
R1-p & 92.89 & 84.67 & 54.67 & 71.78 & \underline{77.33} & 78.44 & 80.00 & \underline{21.67} & \underline{64.10} & \underline{65.00} & 75.67$_{\pm0.78}$ & -3.45 \\
LLM-RM & 92.44 & 83.56 & 58.73 & 71.78 & 75.73 & 77.00 & 81.06 & 25.00 & 56.41 & 67.50 & 75.72$_{\pm0.89}$ & -3.40 \\
kNN & 91.78 & 81.56 & 64.40 & 72.22 & 73.48 & 78.82 & 80.56 & 21.67 & 56.41 & 66.25 & 75.97$_{\pm0.65}$ & -3.15 \\
Label Propagation & 92.67 & 85.78 & 58.00 & 71.78 & 74.22 & 80.00 & 80.22 & 23.33 & 58.97 & 62.50 & 75.97$_{\pm0.72}$ & -3.15 \\
\rowcolor{violet!10} \textbf{MemReward (Ours)} & 92.89 & \underline{86.44} & \textbf{61.11} & \underline{72.00} & 74.44 & \underline{81.78} & \underline{80.44} & \textbf{30.00} & 61.54 & 63.75 & \underline{77.02}$_{\pm0.67}$ & -2.10 \\
R1-Oracle & 92.89 & \textbf{90.22} & \underline{60.44} & \textbf{72.22} & \textbf{79.11} & \textbf{83.11} & \textbf{84.00} & \textbf{30.00} & \textbf{71.79} & \textbf{73.75} & \textbf{79.12}$_{\pm0.46}$ & 0 \\
\bottomrule
\end{tabular}
}
\vspace{-3mm}
\end{table}

\subsection{MemReward Approaches Oracle Performance with Only 20\% Labels}
\label{sec:main_results}
Here, labeled and unlabeled queries are random splits of the same datasets, so this setting tests within-distribution reward propagation; Sections~\ref{sec:ood} and~\ref{sec:generalization} evaluate held-out datasets and cross-domain transfer. Table~\ref{tab:main_results} presents the performance of 10 benchmarks on two model scales. In Qwen2.5-3B, MemReward achieves a mean score of 77.02\%, reaching 97.3\% of Oracle performance while outperforming R1-p by 1.35 points. In Qwen2.5-1.5B, MemReward improves over R1-p by 5.38 points, reaching 96.6\% of Oracle. On both scales, MemReward reduces the gap between partial labels and full supervision, showing that graph-based reward propagation provides an effective training signal regardless of model capacity (comparative case studies in Appendix~\ref{appendix:comparative_cases}). Across six random seeds, MemReward's gains over R1-p are statistically significant on both model sizes, in-domain and out-of-domain (two-tailed paired t-test, $p<0.05$; Appendix~\ref{appendix:stats}). R1-Oracle is a fully supervised reference rather than a per-benchmark upper bound: MemReward exceeds it on several benchmarks, notably GSM8K and GSM-Symbolic for 1.5B and MATH for 3B, where GNN reward predictions are most accurate (Appendix~\ref{appendix:oracle}).

\textbf{MemReward outperforms all reward prediction baselines.} On Qwen2.5-3B, MemReward outperforms LLM-RM by +1.30 and kNN by +1.05. On Qwen2.5-1.5B, the advantage widens: MemReward exceeds LLM-RM by +4.81 and kNN by +3.61, consistent with graph-based reward propagation being more valuable when the policy is weaker. At the reward prediction level, the GNN achieves 83.8\% test-set accuracy, compared with 77.5\% for kNN and 73.5\% for MLP (Appendix~\ref{appendix:gnn_test}). On 3B, Label Propagation reaches 75.97, only 0.30 above R1-p (75.67), and FixMatch-style self-training falls below R1-p at 72.55, whereas MemReward reaches 77.02. The gain thus comes from learned propagation over the graph rather than from adding pseudo-rewards (Appendix~\ref{appendix:ssl}).

\textbf{GNN reward prediction quality.} On held-out validation data, the GNN achieves 0.917 ROC-AUC across all domains, with the highest prediction quality on mathematical reasoning (ROC-AUC 0.936--0.946; see Appendix~\ref{appendix:gnn_quality} for a per-dataset breakdown and Appendix~\ref{appendix:gnn_trace} for propagation traces).

\textbf{Robustness to reward-prediction noise.} Flipping each GNN-predicted reward with probability $p$ (ground-truth rewards untouched; Qwen2.5-1.5B) lowers the average score by 0.72 at $p=30\%$ and 1.01 at $p=50\%$. At $p=70\%$ and $90\%$, where flipped predictions become anti-correlated with the target by construction, the score drops by 10.00 and 21.00 points. Separately, the frozen GNN's accuracy on Qwen2.5-3B stays between 74.9\% and 77.3\% over 410 GRPO steps (Appendix~\ref{appendix:noise}).

\textbf{Mathematical reasoning benefits the most from the predicted rewards} of the GNN; on 1.5B, GSM8K improves by 11.56 points and GSM-Symbolic by 14.89 over R1-p; on 3B, MATH improves by 6.44. This is consistent with MemReward's query-query similarity edges: mathematical queries with similar problem structures share solution strategies, enabling effective reward propagation. Quantitatively, the top-7 nearest neighbors share the same reward label 80\% of the time on average (66--80\% for math, 80--86\% for QA). The sentence encoder only proposes candidate neighbors; the GNN can learn to down-weight reward-inconsistent ones (Appendix~\ref{appendix:gnn_trace}).

\begin{figure}[t]
\centering
\begin{minipage}[t]{0.48\textwidth}
\centering
\includegraphics[width=\linewidth]{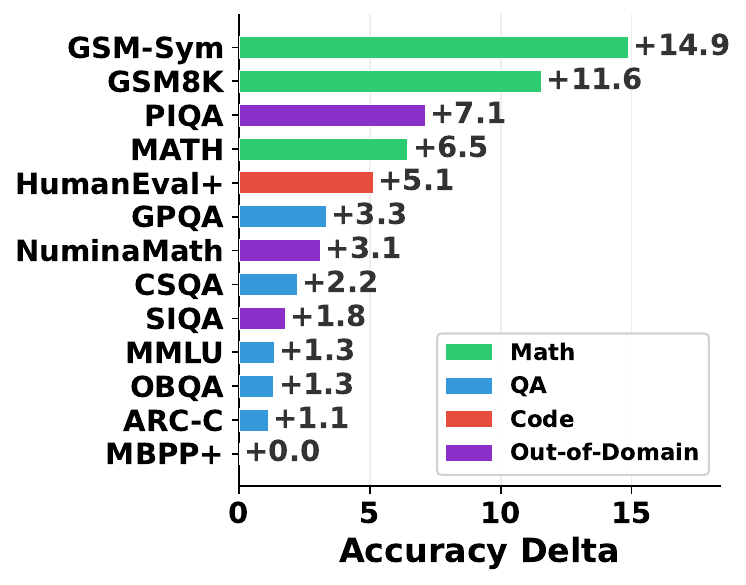}
\captionsetup{justification=raggedright,singlelinecheck=false}
\caption{\textbf{MemReward improves or matches R1-p across all 13 benchmarks on Qwen2.5-1.5B,} with the largest gains on mathematical reasoning (GSM-Sym +14.9, GSM8K +11.6) and the smallest on well-saturated tasks (MBPP+ 0.0).}
\label{fig:benchmark_delta}
\end{minipage}
\hfill
\begin{minipage}[t]{0.48\textwidth}
\centering
\includegraphics[width=\linewidth]{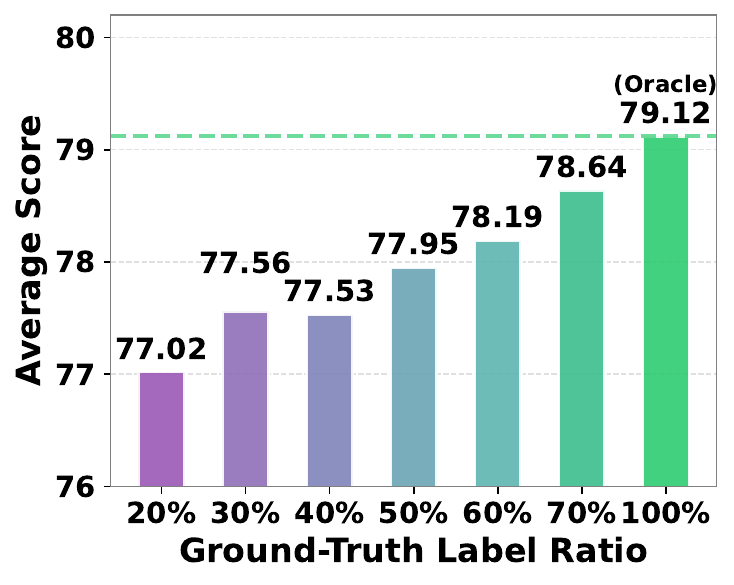}
\caption{\textbf{MemReward performance scales with ground-truth label ratio on Qwen2.5-3B.} Each bar shows the overall average score. Even at 20\% GT, MemReward reaches 97.3\% of Oracle.}
\label{fig:gt_ratio_bar}
\end{minipage}
\end{figure}

\textbf{MemReward scales with label budget.} Figures~\ref{fig:benchmark_delta} and~\ref{fig:gt_ratio_bar} summarize per-benchmark improvements and label ratio scaling. On Qwen2.5-1.5B, MemReward improves or matches R1-p on all 13 benchmarks, with the largest gains in mathematical reasoning (GSM-Sym +14.9, GSM8K +11.6). Performance scales smoothly with the ground-truth label ratio: increasing from 20\% to 70\% labels narrows the gap to Oracle from 2.10 to just 0.48 points (99.4\% of Oracle), with diminishing returns at higher ratios (Appendix~\ref{appendix:gt_ratio}).

\begin{figure}[t]
\centering
\includegraphics[width=\textwidth]{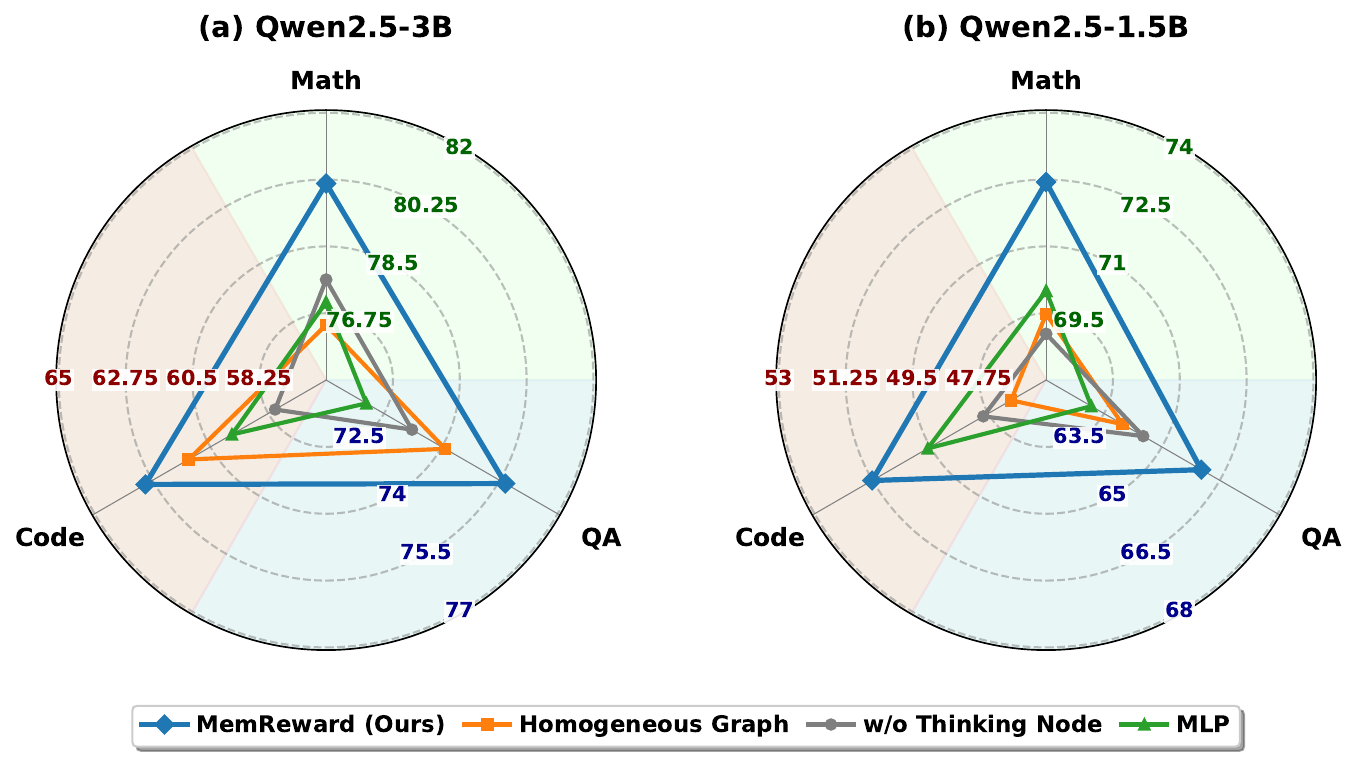}
\caption{\textbf{Ablation studies on (a) Qwen2.5-3B and (b) Qwen2.5-1.5B show each architectural component contributes to performance.} The full model consistently outperforms all ablated variants on both scales across all three task categories.}
\label{fig:ablation_combined}
\vspace{-3mm}
\end{figure}

\subsection{MemReward Approaches Oracle on Out-of-Domain Tasks}
\label{sec:ood}

\begin{table}[t]
\centering
\caption{\textbf{Out-of-domain evaluation. MemReward closely approaches R1-Oracle on out-of-domain tasks using only 20\% ground-truth labels, narrowing the gap to just 0.23 on 3B and 0.75 on 1.5B.} Avg is the mean over six random seeds, with the standard deviation as a subscript.}
\label{tab:ood_results}

\small
\begin{tabular}{lccccc}
\toprule
\textbf{Method} & \textbf{NuminaMath} & \textbf{SIQA} & \textbf{PIQA} & \textbf{Avg} & $\Delta \mathrm{Score}$ \\
\midrule
\multicolumn{6}{l}{\textit{Qwen2.5-1.5B-Instruct}} \\
\midrule[0.3pt]
R1-p & 31.56 & 72.67 & 72.22 & 58.81$_{\pm2.20}$ & -4.75 \\
\rowcolor{violet!10} \textbf{MemReward} & \textbf{34.67} & \underline{74.44} & \underline{79.33} & \underline{62.81}$_{\pm2.05}$ & -0.75 \\
R1-Oracle & \underline{33.78} & \textbf{75.78} & \textbf{81.11} & \textbf{63.56}$_{\pm1.94}$ & 0 \\
\midrule
\multicolumn{6}{l}{\textit{Qwen2.5-3B-Instruct}} \\
\midrule[0.3pt]
R1-p & 36.44 & 74.67 & \textbf{82.22} & 64.44$_{\pm1.05}$ & -2.75 \\
\rowcolor{violet!10} \textbf{MemReward} & \underline{42.22} & \underline{76.89} & 81.78 & \underline{66.96}$_{\pm0.86}$ & -0.23 \\
R1-Oracle & \textbf{42.44} & \textbf{77.11} & \underline{82.00} & \textbf{67.19}$_{\pm0.72}$ & 0 \\
\bottomrule
\end{tabular}
\vspace{-3mm}
\end{table}

\textbf{GNN rewards generalize to out-of-domain tasks.}
Table~\ref{tab:ood_results} presents the results of the out-of-domain evaluation on NuminaMath, SIQA, and PIQA. In this setting, the GNN is trained exclusively on in-domain data (10 benchmarks) during the warmup phase. During online GRPO training, new queries from the held-out tasks are connected to the warmup graph through top-$k$ similarity edges, and the GNN directly predicts rewards without any domain-specific fine-tuning.

MemReward closely approaches Oracle on both model scales: on 3B, the gap narrows to just 0.23 points; on 1.5B, to 0.75 points. On NuminaMath, MemReward closely matches Oracle on both scales, trailing by only 0.22 on 3B and exceeding it by +0.89 on 1.5B, while the remaining benchmarks remain close to Oracle. This shows that the cross-domain graph structure (Section~\ref{sec:cross_domain_graph}), where the GNN aggregates reward signals from semantically similar in-domain experiences through query-query edges, transfers learned reward patterns across domain boundaries. Sensitivity to the query-query connectivity parameter $k$ is evaluated in Appendix~\ref{appendix:topk_ablation}.

\subsection{MemReward Generalizes Across Backbones, Reward Types, and Domains}
\label{sec:generalization}

\begin{table}[t]
\centering
\caption{\textbf{MemReward generalizes beyond Qwen backbones and verifiable rewards.} All settings use ground-truth rewards for only 20\% of the training queries. The TruthfulQA, Chatbot Arena, and HotpotQA rows use Qwen2.5-3B-Instruct; each row reports the metric in parentheses.}
\label{tab:generalization}
\small
\begin{tabular}{llccc}
\toprule
\textbf{Setting (metric)} & \textbf{Reward signal} & \textbf{R1-p} & \textbf{MemReward} & \textbf{R1-Oracle} \\
\midrule
Llama-3.2-3B, 10 benchmarks (Avg) & Verifiable & 67.41 & 70.05 & 72.91 \\
Llama-3.1-8B, 10 benchmarks (Avg) & Verifiable & 74.62 & 77.62 & 79.39 \\
TruthfulQA (truthful \& informative) & LLM judge (binary) & 36.73 & 39.59 & 44.90 \\
Chatbot Arena (preference acc.) & Human preference & 57.33 & 59.00 & 63.00 \\
HotpotQA (token F1) & Continuous (F1) & 39.42 & 42.10 & 47.17 \\
\bottomrule
\end{tabular}
\vspace{-3mm}
\end{table}

Table~\ref{tab:generalization} evaluates MemReward beyond the Qwen backbones and verifiable rewards, keeping the 20\% label budget. With Llama-3.2-3B-Instruct and Llama-3.1-8B-Instruct, MemReward improves over R1-p by 2.64 and 3.00 points, reaching 96.1\% and 97.8\% of R1-Oracle. With Qwen2.5-3B-Instruct, it also improves over R1-p when rewards come from an LLM judge on TruthfulQA~\citep{lin2022truthfulqa} (+2.86), from human pairwise preferences on Chatbot Arena~\citep{chiang2024chatbot} (+1.67), and from continuous token-level F1 on HotpotQA~\citep{yang2018hotpotqa} (+2.68), where the GNN is trained as a regressor. Finally, a GNN trained only on QA labels and used for RL on math raises the math average from 77.36 to 80.79, and the reverse direction raises the QA average from 73.95 to 76.80 (Appendix~\ref{appendix:cross_domain}).

\subsection{Each Architectural Component Contributes to Reward Prediction}
\label{sec:ablation}

We ablate three key architectural components of MemReward by comparing against three variants: (1) \textbf{MLP}, which replaces the GNN with a multi-layer perceptron that predicts rewards from query embeddings alone, without graph structure; (2) \textbf{Homogeneous Graph}, which removes edge type distinctions and treats all nodes uniformly; (3) \textbf{w/o Thinking Node}, which removes the intermediate thinking nodes from the heterogeneous graph.

Figure~\ref{fig:ablation_combined} presents results per-category on both model scales, where the category-level accuracy is computed by averaging over all benchmarks within each task category. On 3B, the full model achieves 80.1\% in Math, 75.6\% in QA, and 63.0\% in Code, consistently outperforming all ablated variants. On 1.5B, the full model achieves 72.4\% in Math, 66.0\% in QA, and 51.3\% in Code.

\textbf{Graph structure captures inter-query reward patterns.} The MLP baseline, which predicts rewards from query embeddings alone without graph structure, suffers the largest drop on QA: 72.0\% on 3B (a gap of 3.6 points) and 63.2\% on 1.5B (a gap of 2.8 points). QA queries span diverse topics, including science, social reasoning, and physical commonsense, where individual embeddings provide limited reward signals. This shows that the edges of query-query similarity (Section~\ref{sec:cross_domain_graph}) capture exploitable structural dependencies between experiences, addressing the challenge identified in Section~\ref{sec:intro}.

\textbf{Heterogeneous edge types benefit structured reasoning.} The homogeneous graph variant, which treats all edges uniformly, retains 74.1\% on QA but drops to 76.4\% on Math, a gap of 3.7 points from the full model. Math problems exhibit distinct structural relationships at different levels: query-query edges capture problem similarity (e.g., two quadratic equations), while query-thinking and thinking-answer edges capture solution correctness. Collapsing these into a single edge type loses this hierarchical signal, demonstrating that the heterogeneous edge design (Section~\ref{sec:cross_domain_graph}) preserves task-specific structural relationships essential for accurate reward prediction.

\textbf{Thinking nodes benefit multi-step reasoning tasks.} Removing thinking nodes degrades Code to 58.0\% on 3B (a drop of 5.0 points) and Math to 77.6\% (a drop of 2.5 points). In math and code, the thinking process encodes intermediate reasoning steps (e.g., equation manipulations, algorithmic logic) that determine whether the final answer is correct. Without thinking nodes, the GNN can only compare queries and answers, losing the fine-grained signal about \emph{how} the model reasons. On 1.5B, the w/o Thinking variant similarly falls to 69.0\% in Math (a gap of 3.4 points). The consistency of component contributions across both model scales shows that the heterogeneous three-node design (Section~\ref{sec:cross_domain_graph}) captures reasoning structure that generalizes across model capacities. We also evaluate reward threshold sensitivity in Appendix~\ref{appendix:threshold_ablation}.

\section{Related Work}

\textbf{Reinforcement Learning for LLMs.} Early work established the RLHF paradigm~\citep{ouyang2022training}, training reward models from human evaluations and optimizing via PPO~\citep{schulman2017proximal}. Subsequent efforts reduced pipeline complexity through AI feedback~\citep{lee2023rlaif} and direct preference optimization~\citep{rafailov2023direct}. Process-level supervision further improved reward quality via step-by-step verification~\citep{lightman2023let} and automated labeling~\citep{wang2024math}. More recently, DeepSeek-R1~\citep{guo2025deepseek} used GRPO with group-based advantage estimation for reasoning. These methods assume full reward supervision, whereas MemReward addresses limited labels. Recent work further addresses reward model reliability: ensemble-based uncertainty penalization mitigates overoptimization~\citep{coste2023reward}, and noise-corrected gradient estimation handles noisy verifiers~\citep{mansouri2025noise}. MemReward is complementary, addressing the orthogonal challenge of label scarcity rather than reward model calibration.

\textbf{Graph Neural Networks for Language Tasks.} GNNs model relational structures through iterative message passing~\citep{gilmer2017neural}, with foundational architectures including GCN~\citep{kipf2016semi}, GAT~\citep{velivckovic2017graph}, and GATv2~\citep{brody2021attentive}. Heterogeneous extensions such as HAN~\citep{wang2019heterogeneous} and R-GCN~\citep{schlichtkrull2018modeling} handle multiple node and edge types. Our work extends heterogeneous GNNs to reward prediction, constructing graphs over queries, thinking processes, and answers to propagate reward labels.

\textbf{Semi-supervised Learning and Label Propagation.} Semi-supervised learning addresses label scarcity through graph-based label propagation~\citep{zhu2002learning,iscen2019label} and consistency-based methods such as pseudo-labeling~\citep{lee2013pseudo} and FixMatch~\citep{sohn2020fixmatch}. In the LLM context, Constitutional AI~\citep{bai2022constitutional} uses AI feedback to reduce labeling. Self-training approaches iteratively expand labeled sets using model predictions~\citep{xie2020self}, but risk error accumulation without structural constraints. More broadly, these approaches either rely on fixed similarity metrics or operate on individual samples without exploiting structural dependencies. MemReward bridges these paradigms, predicting rewards through learnable message passing over query-thinking-answer structures. We compare against both label propagation and FixMatch-style self-training in Section~\ref{sec:main_results}.

\section{Conclusion}

We proposed MemReward, a graph-based experience memory framework that addresses reward label scarcity in reinforcement learning for LLMs. By organizing queries, thinking processes, and answers into a heterogeneous graph, MemReward propagates rewards from a small set of labeled rollouts to unlabeled ones during online policy optimization. Across 13 benchmarks spanning mathematics, question answering, and code generation on two Qwen2.5 model scales, MemReward achieves near-Oracle performance with only 20\% ground-truth labels and closely approaches Oracle on out-of-domain tasks; the gains carry over to Llama backbones up to 8B and to LLM-judge, human-preference, and continuous rewards. Ablation studies confirm that each architectural component contributes to prediction quality. Overall, MemReward demonstrates that graph-based experience memory is an effective and scalable approach for enabling label-efficient RL fine-tuning of LLMs. We discuss limitations and future directions in Appendix~\ref{appendix:limitations}.

\newpage

\begin{ack}
We sincerely appreciate the support from the research gift from Meta.
\end{ack}

\bibliography{references}

@inproceedings{qiu2025reward,
  title={Reward generalization in rlhf: A topological perspective},
  author={Qiu, Tianyi Alex and Zeng, Fanzhi and Ji, Jiaming and Yan, Dong and Wang, Kaile and Zhou, Jiayi and Han, Yang and Dai, Josef and Pan, Xuehai and Yang, Yaodong},
  booktitle={Findings of the Association for Computational Linguistics: ACL 2025},
  pages={15884--15930},
  year={2025}
}

@article{huang2021combining,
  title={Combining label propagation and simple models out-performs graph neural networks},
  author={Huang, Qian and He, Horace and Singh, Abhay and Lim, Ser-Nam and Benson, Austin R},
  journal={arXiv preprint arXiv:2010.13993},
  year={2020}
}

@article{guo2025deepseek,
  title={Deepseek-r1: Incentivizing reasoning capability in llms via reinforcement learning},
  author={Guo, Daya and Yang, Dejian and Zhang, Haowei and Song, Junxiao and Wang, Peiyi and Zhu, Qihao and Xu, Runxin and Zhang, Ruoyu and Ma, Shirong and Bi, Xiao and others},
  journal={arXiv preprint arXiv:2501.12948},
  year={2025}
}

@article{ouyang2022training,
  title={Training language models to follow instructions with human feedback},
  author={Ouyang, Long and Wu, Jeffrey and Jiang, Xu and Almeida, Diogo and Wainwright, Carroll and Mishkin, Pamela and Zhang, Chong and Agarwal, Sandhini and Slama, Katarina and Ray, Alex and others},
  journal={Advances in neural information processing systems},
  volume={35},
  pages={27730--27744},
  year={2022}
}

@article{schulman2017proximal,
  title={Proximal policy optimization algorithms},
  author={Schulman, John and Wolski, Filip and Dhariwal, Prafulla and Radford, Alec and Klimov, Oleg},
  journal={arXiv preprint arXiv:1707.06347},
  year={2017}
}

@article{cobbe2021training,
  title={Training verifiers to solve math word problems},
  author={Cobbe, Karl and Kosaraju, Vineet and Bavarian, Mohammad and Chen, Mark and Jun, Heewoo and Kaiser, Lukasz and Plappert, Matthias and Tworek, Jerry and Hilton, Jacob and Nakano, Reiichiro and others},
  journal={arXiv preprint arXiv:2110.14168},
  year={2021}
}

@article{hendrycks2021measuring,
  title={Measuring mathematical problem solving with the math dataset},
  author={Hendrycks, Dan and Burns, Collin and Kadavath, Saurav and Arora, Akul and Basart, Steven and Tang, Eric and Song, Dawn and Steinhardt, Jacob},
  journal={arXiv preprint arXiv:2103.03874},
  year={2021}
}

@article{li2024numinamath,
  title={Numinamath: The largest public dataset in ai4maths with 860k pairs of competition math problems and solutions},
  author={Li, Jia and Beeching, Edward and Tunstall, Lewis and Lipkin, Ben and Soletskyi, Roman and Huang, Shengyi and Rasul, Kashif and Yu, Longhui and Jiang, Albert Q and Shen, Ziju and others},
  journal={Hugging Face repository},
  volume={13},
  number={9},
  pages={9},
  year={2024}
}

@article{austin2021program,
  title={Program synthesis with large language models},
  author={Austin, Jacob and Odena, Augustus and Nye, Maxwell and Bosma, Maarten and Michalewski, Henryk and Dohan, David and Jiang, Ellen and Cai, Carrie and Terry, Michael and Le, Quoc and others},
  journal={arXiv preprint arXiv:2108.07732},
  year={2021}
}

@article{chen2021evaluating,
  title={Evaluating large language models trained on code},
  author={Chen, Mark and Tworek, Jerry and Jun, Heewoo and Yuan, Qiming and Pinto, Henrique Ponde De Oliveira and Kaplan, Jared and Edwards, Harri and Burda, Yuri and Joseph, Nicholas and Brockman, Greg and others},
  journal={arXiv preprint arXiv:2107.03374},
  year={2021}
}

@article{hendrycks2020measuring,
  title={Measuring massive multitask language understanding},
  author={Hendrycks, Dan and Burns, Collin and Basart, Steven and Zou, Andy and Mazeika, Mantas and Song, Dawn and Steinhardt, Jacob},
  journal={arXiv preprint arXiv:2009.03300},
  year={2020}
}

@inproceedings{talmor2019commonsenseqa,
  title={Commonsenseqa: A question answering challenge targeting commonsense knowledge},
  author={Talmor, Alon and Herzig, Jonathan and Lourie, Nicholas and Berant, Jonathan},
  booktitle={Proceedings of the 2019 Conference of the North American Chapter of the Association for Computational Linguistics: Human Language Technologies, Volume 1 (Long and Short Papers)},
  pages={4149--4158},
  year={2019}
}

@inproceedings{mihaylov2018can,
  title={Can a suit of armor conduct electricity? a new dataset for open book question answering},
  author={Mihaylov, Todor and Clark, Peter and Khot, Tushar and Sabharwal, Ashish},
  booktitle={Proceedings of the 2018 conference on empirical methods in natural language processing},
  pages={2381--2391},
  year={2018}
}

@article{clark2018think,
  title={Think you have solved question answering? try arc, the ai2 reasoning challenge},
  author={Clark, Peter and Cowhey, Isaac and Etzioni, Oren and Khot, Tushar and Sabharwal, Ashish and Schoenick, Carissa and Tafjord, Oyvind},
  journal={arXiv preprint arXiv:1803.05457},
  year={2018}
}

@inproceedings{rein2024gpqa,
  title={Gpqa: A graduate-level google-proof q\&a benchmark},
  author={Rein, David and Hou, Betty Li and Stickland, Asa Cooper and Petty, Jackson and Pang, Richard Yuanzhe and Dirani, Julien and Michael, Julian and Bowman, Samuel R},
  booktitle={First conference on language modeling},
  year={2024}
}

@inproceedings{sap2019socialiqa,
  title={Social IQa: Commonsense reasoning about social interactions},
  author={Sap, Maarten and Rashkin, Hannah and Chen, Derek and Le Bras, Ronan and Choi, Yejin},
  booktitle={Proceedings of the 2019 conference on empirical methods in natural language processing and the 9th international joint conference on natural language processing (EMNLP-IJCNLP)},
  pages={4463--4473},
  year={2019}
}

@inproceedings{bisk2020piqa,
  title={Piqa: Reasoning about physical commonsense in natural language},
  author={Bisk, Yonatan and Zellers, Rowan and Gao, Jianfeng and Choi, Yejin and others},
  booktitle={Proceedings of the AAAI conference on artificial intelligence},
  volume={34},
  number={05},
  pages={7432--7439},
  year={2020}
}

@article{lee2023rlaif,
  title={Rlaif vs. rlhf: Scaling reinforcement learning from human feedback with ai feedback},
  author={Lee, Harrison and Phatale, Samrat and Mansoor, Hassan and Mesnard, Thomas and Ferret, Johan and Lu, Kellie and Bishop, Colton and Hall, Ethan and Carbune, Victor and Rastogi, Abhinav and others},
  journal={arXiv preprint arXiv:2309.00267},
  year={2023}
}

@article{kipf2016semi,
  title={Semi-supervised classification with graph convolutional networks},
  author={Kipf, Thomas N and Welling, Max},
  journal={arXiv preprint arXiv:1609.02907},
  year={2016}
}

@article{velivckovic2017graph,
  title={Graph attention networks},
  author={Veli{\v{c}}kovi{\'c}, Petar and Cucurull, Guillem and Casanova, Arantxa and Romero, Adriana and Lio, Pietro and Bengio, Yoshua},
  journal={arXiv preprint arXiv:1710.10903},
  year={2017}
}

@inproceedings{wang2019heterogeneous,
  title={Heterogeneous graph attention network},
  author={Wang, Xiao and Ji, Houye and Shi, Chuan and Wang, Bai and Ye, Yanfang and Cui, Peng and Yu, Philip S},
  booktitle={The world wide web conference},
  pages={2022--2032},
  year={2019}
}

@inproceedings{schlichtkrull2018modeling,
  title={Modeling relational data with graph convolutional networks},
  author={Schlichtkrull, Michael and Kipf, Thomas N and Bloem, Peter and Van Den Berg, Rianne and Titov, Ivan and Welling, Max},
  booktitle={European semantic web conference},
  pages={593--607},
  year={2018},
  organization={Springer}
}

@inproceedings{gilmer2017neural,
  title={Neural message passing for quantum chemistry},
  author={Gilmer, Justin and Schoenholz, Samuel S and Riley, Patrick F and Vinyals, Oriol and Dahl, George E},
  booktitle={International conference on machine learning},
  pages={1263--1272},
  year={2017},
  organization={Pmlr}
}

@article{rafailov2023direct,
  title={Direct preference optimization: Your language model is secretly a reward model},
  author={Rafailov, Rafael and Sharma, Archit and Mitchell, Eric and Manning, Christopher D and Ermon, Stefano and Finn, Chelsea},
  journal={Advances in neural information processing systems},
  volume={36},
  pages={53728--53741},
  year={2023}
}

@article{bai2022constitutional,
  title={Constitutional ai: Harmlessness from ai feedback},
  author={Bai, Yuntao and Kadavath, Saurav and Kundu, Sandipan and Askell, Amanda and Kernion, Jackson and Jones, Andy and Chen, Anna and Goldie, Anna and Mirhoseini, Azalia and McKinnon, Cameron and others},
  journal={arXiv preprint arXiv:2212.08073},
  year={2022}
}

@techreport{zhu2002learning,
  title={Learning from labeled and unlabeled data with label propagation},
  author={Zhu, Xiaojin and Ghahramani, Zoubin},
  institution={Carnegie Mellon University},
  number={CMU-CALD-02-107},
  year={2002}
}

@inproceedings{lee2013pseudo,
  title={Pseudo-label: The simple and efficient semi-supervised learning method for deep neural networks},
  author={Lee, Dong-Hyun and others},
  booktitle={Workshop on challenges in representation learning, ICML},
  volume={3},
  number={2},
  pages={896},
  year={2013},
  organization={Atlanta}
}

@inproceedings{lightman2023let,
  title={Let's verify step by step},
  author={Lightman, Hunter and Kosaraju, Vineet and Burda, Yuri and Edwards, Harrison and Baker, Bowen and Lee, Teddy and Leike, Jan and Schulman, John and Sutskever, Ilya and Cobbe, Karl},
  booktitle={The twelfth international conference on learning representations},
  year={2023}
}

@inproceedings{iscen2019label,
  title={Label propagation for deep semi-supervised learning},
  author={Iscen, Ahmet and Tolias, Giorgos and Avrithis, Yannis and Chum, Ondrej},
  booktitle={Proceedings of the IEEE/CVF conference on computer vision and pattern recognition},
  pages={5070--5079},
  year={2019}
}

@article{sohn2020fixmatch,
  title={Fixmatch: Simplifying semi-supervised learning with consistency and confidence},
  author={Sohn, Kihyuk and Berthelot, David and Carlini, Nicholas and Zhang, Zizhao and Zhang, Han and Raffel, Colin A and Cubuk, Ekin Dogus and Kurakin, Alexey and Li, Chun-Liang},
  journal={Advances in neural information processing systems},
  volume={33},
  pages={596--608},
  year={2020}
}

@article{yu2025dapo,
  title={Dapo: An open-source llm reinforcement learning system at scale},
  author={Yu, Qiying and Zhang, Zheng and Zhu, Ruofei and Yuan, Yufeng and Zuo, Xiaochen and Yue, Yu and Dai, Weinan and Fan, Tiantian and Liu, Gaohong and Liu, Lingjun and others},
  journal={arXiv preprint arXiv:2503.14476},
  year={2025}
}

@inproceedings{wang2024math,
  title={Math-shepherd: Verify and reinforce llms step-by-step without human annotations},
  author={Wang, Peiyi and Li, Lei and Shao, Zhihong and Xu, Runxin and Dai, Damai and Li, Yifei and Chen, Deli and Wu, Yu and Sui, Zhifang},
  booktitle={Proceedings of the 62nd Annual Meeting of the Association for Computational Linguistics (Volume 1: Long Papers)},
  pages={9426--9439},
  year={2024}
}

@article{brody2021attentive,
  title={How attentive are graph attention networks?},
  author={Brody, Shaked and Alon, Uri and Yahav, Eran},
  journal={arXiv preprint arXiv:2105.14491},
  year={2021}
}

@article{mirzadeh2024gsm,
  title={Gsm-symbolic: Understanding the limitations of mathematical reasoning in large language models},
  author={Mirzadeh, Iman and Alizadeh, Keivan and Shahrokhi, Hooman and Tuzel, Oncel and Bengio, Samy and Farajtabar, Mehrdad},
  journal={arXiv preprint arXiv:2410.05229},
  year={2024}
}

@article{liu2023your,
  title={Is your code generated by chatgpt really correct? rigorous evaluation of large language models for code generation},
  author={Liu, Jiawei and Xia, Chunqiu Steven and Wang, Yuyao and Zhang, Lingming},
  journal={Advances in neural information processing systems},
  volume={36},
  pages={21558--21572},
  year={2023}
}

@inproceedings{reimers2019sentence,
  title={Sentence-bert: Sentence embeddings using siamese bert-networks},
  author={Reimers, Nils and Gurevych, Iryna},
  booktitle={Proceedings of the 2019 conference on empirical methods in natural language processing and the 9th international joint conference on natural language processing (EMNLP-IJCNLP)},
  pages={3982--3992},
  year={2019}
}

@article{fey2019fast,
  title={Fast graph representation learning with PyTorch Geometric},
  author={Fey, Matthias and Lenssen, Jan Eric},
  journal={arXiv preprint arXiv:1903.02428},
  year={2019}
}

@article{cui2025process,
  title={Process reinforcement through implicit rewards},
  author={Cui, Ganqu and Yuan, Lifan and Wang, Zefan and Wang, Hanbin and Zhang, Yuchen and Chen, Jiacheng and Li, Wendi and He, Bingxiang and Fan, Yuchen and Yu, Tianyu and others},
  journal={arXiv preprint arXiv:2502.01456},
  year={2025}
}

@article{jin2025search,
  title={Search-r1: Training llms to reason and leverage search engines with reinforcement learning},
  author={Jin, Bowen and Zeng, Hansi and Yue, Zhenrui and Yoon, Jinsung and Arik, Sercan and Wang, Dong and Zamani, Hamed and Han, Jiawei},
  journal={arXiv preprint arXiv:2503.09516},
  year={2025}
}

@article{coste2023reward,
  title={Reward model ensembles help mitigate overoptimization},
  author={Coste, Thomas and Anwar, Usman and Kirk, Robert and Krueger, David},
  journal={arXiv preprint arXiv:2310.02743},
  year={2023}
}

@article{mansouri2025noise,
  title={Noise-corrected GRPO: From Noisy Rewards to Unbiased Gradients},
  author={Mansouri, Omar El and Seddik, Mohamed El Amine and Lahlou, Salem},
  journal={arXiv preprint arXiv:2510.18924},
  year={2025}
}

@inproceedings{xie2020self,
  title={Self-training with noisy student improves imagenet classification},
  author={Xie, Qizhe and Luong, Minh-Thang and Hovy, Eduard and Le, Quoc V},
  booktitle={Proceedings of the IEEE/CVF conference on computer vision and pattern recognition},
  pages={10687--10698},
  year={2020}
}

@inproceedings{lin2022truthfulqa,
  title={{TruthfulQA}: Measuring How Models Mimic Human Falsehoods},
  author={Lin, Stephanie and Hilton, Jacob and Evans, Owain},
  booktitle={Proceedings of the 60th Annual Meeting of the Association for Computational Linguistics (Volume 1: Long Papers)},
  pages={3214--3252},
  year={2022}
}

@inproceedings{yang2018hotpotqa,
  title={{HotpotQA}: A Dataset for Diverse, Explainable Multi-hop Question Answering},
  author={Yang, Zhilin and Qi, Peng and Zhang, Saizheng and Bengio, Yoshua and Cohen, William W. and Salakhutdinov, Ruslan and Manning, Christopher D.},
  booktitle={Proceedings of the 2018 Conference on Empirical Methods in Natural Language Processing},
  pages={2369--2380},
  year={2018}
}

@article{grattafiori2024llama,
  title={The {Llama} 3 Herd of Models},
  author={Grattafiori, Aaron and Dubey, Abhimanyu and Jauhri, Abhinav and Pandey, Abhinav and Kadian, Abhishek and others},
  journal={arXiv preprint arXiv:2407.21783},
  year={2024}
}

@inproceedings{zheng2023judging,
  title={Judging {LLM}-as-a-Judge with {MT-Bench} and {Chatbot Arena}},
  author={Zheng, Lianmin and Chiang, Wei-Lin and Sheng, Ying and Zhuang, Siyuan and Wu, Zhanghao and Zhuang, Yonghao and Lin, Zi and Li, Zhuohan and Li, Dacheng and Xing, Eric P. and Zhang, Hao and Gonzalez, Joseph E. and Stoica, Ion},
  booktitle={Advances in Neural Information Processing Systems (Datasets and Benchmarks Track)},
  volume={36},
  pages={46595--46623},
  year={2023}
}

@article{chiang2024chatbot,
  title={{Chatbot Arena}: An Open Platform for Evaluating {LLMs} by Human Preference},
  author={Chiang, Wei-Lin and Zheng, Lianmin and Sheng, Ying and Angelopoulos, Anastasios Nikolas and Li, Tianle and Li, Dacheng and Zhang, Hao and Zhu, Banghua and Jordan, Michael and Gonzalez, Joseph E. and others},
  journal={arXiv preprint arXiv:2403.04132},
  year={2024}
}

@article{xu2026deepseek,
  title={{DeepSeek-V4}: Towards Highly Efficient Million-Token Context Intelligence},
  author={Xu, Anyi and Lin, Bangcai and Xue, Bing and Wang, Bingxuan and Xu, Bingzheng and Wu, Bochao and Zhang, Bowei and Lin, Chaofan and Dong, Chen and Ling, Chenchen and others},
  journal={arXiv preprint arXiv:2606.19348},
  year={2026}
}
\bibliographystyle{plain}

\newpage
\DoToC
\appendix
\section{Limitations, Future Work, and Broader Impact}
\label{appendix:limitations}

\textbf{Limitations.} We evaluate verifiable rewards on 13 benchmarks with Qwen2.5-1.5B and 3B, and on 10 benchmarks with Llama-3.2-3B and Llama-3.1-8B. Additional experiments with Qwen2.5-3B use LLM-judge (TruthfulQA), human-preference (Chatbot Arena), and continuous token-F1 (HotpotQA) rewards. Two design choices define the current scope. (1) The GNN is frozen after warmup, which avoids a moving-target feedback loop between the reward predictor and the policy, while the 20\% ground-truth rewards continuously anchor optimization. Its accuracy stays stable over 410 GRPO steps, and training remains robust when up to half of the predicted rewards are randomly flipped (Appendix~\ref{appendix:noise}). This study characterizes robustness to random prediction errors; systematic prediction bias and policy-dependent reward exploitation lie outside it. (2) Query-query edges are built with a fixed general-purpose sentence encoder (all-MiniLM-L6-v2) that is not trained for reward relevance. Its neighborhoods nonetheless carry reward-relevant signal: the top-7 nearest queries share the reward label about 80\% of the time, and a GNN trained on one domain predicts rewards in another 12.95 (QA$\rightarrow$Math) and 7.35 (Math$\rightarrow$QA) points more accurately than a graph-free MLP (Appendix~\ref{appendix:cross_domain}).

\textbf{Future Work.} Natural next steps include extending MemReward to further forms of supervision, such as graded expert assessment of mathematical proofs and multi-level human ratings, and to larger backbones; characterizing systematic reward bias and policy-dependent reward exploitation, together with safeguards against them; learning reward-aware similarity jointly with the GNN; and exploring richer graph topologies, such as directed or hierarchical experience graphs that order experiences by generation time or label confidence. Scaling graph construction to larger rollout pools is another promising avenue.

\textbf{Broader Impact.} By reducing the reliance on costly human labels through graph-based reward propagation, MemReward has the potential to democratize access to RL-based LLM training for researchers with limited labeling budgets. We do not foresee specific negative societal consequences beyond those generally associated with improving LLM reasoning capabilities. As with all advances in LLM performance, downstream applications should be deployed responsibly with appropriate safeguards.

\section{Ground-Truth Label Ratio Analysis}
\label{appendix:gt_ratio}

Table~\ref{tab:gt_ratio} examines how MemReward performs as the proportion of ground-truth labels varies from 20\% to 70\% on Qwen2.5-3B, with the remaining queries receiving GNN-predicted rewards (see also Figure~\ref{fig:gt_ratio_bar} in the main text). Even at the lowest label budget (20\% GT + 80\% GNN), MemReward already achieves 97.3\% of Oracle performance (77.02 vs.\ 79.12). As the GT ratio increases to 60\%, MemReward reaches 98.8\% of Oracle (78.19 vs.\ 79.12), and at 70\% the gap narrows to just 0.48 points (78.64 vs.\ 79.12), reaching 99.4\% of Oracle.

\begin{table}[t]
\centering
\caption{\textbf{MemReward maintains near-Oracle performance across ground-truth label ratios on Qwen2.5-3B.} We vary the proportion of ground-truth labels from 20\% to 70\%, with the remaining queries receiving GNN-predicted rewards. Avg is the mean over six random seeds, with the standard deviation as a subscript.}
\label{tab:gt_ratio}

\small
\resizebox{1.0\textwidth}{!}{
\begin{tabular}{lcccccccccccc}
\toprule
\multirow{2}{*}{\textbf{GT Ratio}} & \multicolumn{3}{c}{\textbf{Math}} & \multicolumn{5}{c}{\textbf{QA}} & \multicolumn{2}{c}{\textbf{Code}} & \multirow{2}{*}{\textbf{Avg}} & \multirow{2}{*}{$\Delta \mathrm{Score}$} \\
\cmidrule(lr){2-4} \cmidrule(lr){5-9} \cmidrule(lr){10-11}
& GSM8K & GSM-sym & MATH & MMLU & CSQA & OBQA & ARC-C & GPQA & HumanEval+ & MBPP+ & & \\
\midrule
20\% & 92.89 & 86.44 & 61.11 & 72.00 & 74.44 & 81.78 & 80.44 & \textbf{30.00} & 61.54 & 63.75 & 77.02$_{\pm0.67}$ & -2.10 \\
30\% & \textbf{93.56} & 90.67 & 61.78 & 69.56 & 78.22 & 79.56 & 81.11 & 25.00 & 56.41 & 62.50 & 77.56$_{\pm0.63}$ & -1.56 \\
40\% & 92.67 & 91.78 & 57.78 & \textbf{72.44} & 76.22 & 81.56 & \underline{83.11} & 23.33 & 58.97 & 55.00 & 77.53$_{\pm0.56}$ & -1.59 \\
50\% & 92.44 & \textbf{92.89} & \underline{63.56} & 69.11 & 76.22 & 80.44 & 82.22 & 20.00 & 64.10 & 65.00 & 77.95$_{\pm0.53}$ & -1.17 \\
60\% & \underline{93.11} & \underline{92.44} & 56.89 & 70.00 & \underline{78.44} & \underline{82.44} & 82.89 & \underline{28.33} & \underline{66.67} & \underline{71.25} & 78.19$_{\pm0.50}$ & -0.93 \\
70\% & 92.00 & 89.33 & \textbf{66.22} & 72.00 & 78.00 & 82.22 & 81.78 & 20.00 & \textbf{71.79} & 63.75 & \underline{78.64}$_{\pm0.49}$ & -0.48 \\
100\% (Oracle) & 92.89 & 90.22 & 60.44 & \underline{72.22} & \textbf{79.11} & \textbf{83.11} & \textbf{84.00} & \textbf{30.00} & \textbf{71.79} & \textbf{73.75} & \textbf{79.12}$_{\pm0.46}$ & 0 \\
\bottomrule
\end{tabular}
}
\end{table}

\section{Implementation Details}
\label{appendix:hyperparams}

We implement the heterogeneous GNN using PyTorch Geometric~\citep{fey2019fast} with \texttt{HeteroConv} as the heterogeneous message-passing wrapper and \texttt{GATv2Conv}~\citep{brody2021attentive} as the per-relation convolution operator, combined with \texttt{torch\_scatter} for efficient sparse aggregation. The GNN architecture consists of 2 layers with a hidden dimension of 512, 4 attention heads, and a dropout rate of 0.1. For in-domain and out-of-domain experiments, we encode queries, thinking processes, and answers using \texttt{all-MiniLM-L6-v2}~\citep{reimers2019sentence} (384-dim). All embeddings are projected to the hidden dimension via type-specific linear transformations. Query-query edges are constructed using top-$k$ cosine similarity with $k=7$; the kNN baseline uses the same $k=7$ for a fair comparison. The GNN is trained with the Adam optimizer (learning rate $1 \times 10^{-3}$) using binary cross-entropy loss for 150 epochs with early stopping (patience 20).

For GRPO training, we use the veRL framework with the following hyperparameters: actor learning rate $1 \times 10^{-6}$, batch size 128, micro-batch size 32, and maximum response length 1024. We enable KL loss regularization ($\beta = 1 \times 10^{-3}$) and advantage clipping ($\epsilon = 0.2$). Response generation uses a temperature of 1.0 for diverse sampling. Each query generates $N=8$ responses for advantage estimation. Training runs for 410 steps, with model selection based on validation performance. The Llama-3.2-3B-Instruct and Llama-3.1-8B-Instruct experiments use the same settings. All experiments are conducted on NVIDIA RTX PRO 6000 GPUs with BF16 precision.
\begin{table}[t]
\centering
\caption{\textbf{GRPO training hyperparameters.} All RL fine-tuning experiments use these settings with Qwen2.5-1.5B-Instruct, Qwen2.5-3B-Instruct, Llama-3.2-3B-Instruct, and Llama-3.1-8B-Instruct as the backbone models.}
\label{tab:hyperparams}

\small
\begin{tabular}{lc}
\toprule
\textbf{Hyperparameter} & \textbf{Value} \\
\midrule
Actor learning rate & $1e{-}6$ \\
train\_batch\_size & 128 \\
mini\_batch\_size & 128 \\
micro\_batch\_size & 32 \\
Training steps & 410 \\
Max response length & 1024 \\
Num of rollouts & 8 \\
Rollout temp ($\tau$) & 1.0 \\
KL penalty ($\beta$) & $1e{-}3$ \\
Advantage clip ($\epsilon$) & 0.2 \\
\bottomrule
\end{tabular}
\end{table}

\begin{table}[t]
\centering
\caption{\textbf{GNN hyperparameters.} Configuration of the heterogeneous graph neural network used for reward prediction.}
\label{tab:gnn_hyperparams}

\small
\begin{tabular}{lc}
\toprule
\textbf{Hyperparameter} & \textbf{Value} \\
\midrule
Input dimension & 384 \\
Hidden dimension & 512 \\
Number of GNN layers & 2 \\
Number of attention heads & 4 \\
Dropout & 0.1 \\
Learning rate & $1e{-}3$ \\
Training epochs & 150 \\
Query-Query top-$k$ & 7 \\
\bottomrule
\end{tabular}
\end{table}

\section{Dataset Statistics}
\label{appendix:dataset_statistics}

Table~\ref{tab:dataset_standard} summarizes the data splits used across all in-domain benchmarks.

\begin{table}[t]
\centering
\caption{\textbf{Dataset statistics for in-domain evaluation.} For each dataset, we randomly sample 1,500 instances and split them into training, validation, and testing sets. R1-Oracle uses all training queries with 100\% ground-truth labels; MemReward uses all training queries with 20\% GT labels and 80\% GNN-predicted rewards; R1-p uses only the 20\% labeled rollouts and discards the rest. Validation labels are available in all settings for model selection. Note that GPQA, HumanEval+, and MBPP+ have fewer than 1,500 available instances due to their original dataset sizes.}
\label{tab:dataset_standard}

\small
\resizebox{1.0\textwidth}{!}{
\begin{tabular}{lcccccccccc}
\toprule
\multirow{2}{*}{\textbf{Split}} & \multicolumn{3}{c}{\textbf{Math}} & \multicolumn{5}{c}{\textbf{QA}} & \multicolumn{2}{c}{\textbf{Code}} \\
\cmidrule(lr){2-4} \cmidrule(lr){5-9} \cmidrule(lr){10-11}
& GSM8K & GSM-sym & MATH & MMLU & CSQA & OBQA & ARC-C & GPQA & HumanEval+ & MBPP+ \\
\midrule
R1-Oracle & 750 & 750 & 750 & 750 & 750 & 750 & 750 & 99 & 65 & 132 \\
MemReward & 750 & 750 & 750 & 750 & 750 & 750 & 750 & 99 & 65 & 132 \\
R1-p & 150 & 150 & 150 & 150 & 150 & 150 & 150 & 19 & 13 & 26 \\
\midrule
Validation & 300 & 300 & 300 & 300 & 300 & 300 & 300 & 39 & 26 & 52 \\
Testing & 450 & 450 & 450 & 450 & 450 & 450 & 450 & 60 & 39 & 80 \\
\bottomrule
\end{tabular}
}
\end{table}

\section{Dataset Descriptions}
\label{appendix:dataset_descriptions}

We provide detailed descriptions of all datasets used in our evaluation, organized by task category.

\subsection{Mathematical Reasoning}

\textbf{GSM8K}~\citep{cobbe2021training} is a dataset of 8.5K grade school math word problems requiring multi-step arithmetic reasoning. Problems involve basic operations (addition, subtraction, multiplication, and division) and require 2-8 reasoning steps to solve. We evaluate using an exact match on the final numerical answer.

\textbf{GSM-Symbolic}~\citep{mirzadeh2024gsm} is a variant of GSM8K generated from symbolic templates that vary names and numerical values while preserving the underlying problem structure. This variant helps assess whether models truly understand mathematical reasoning or merely memorize numerical patterns.

\textbf{MATH}~\citep{hendrycks2021measuring} contains 12.5K challenging competition mathematics problems spanning algebra, geometry, number theory, counting and probability, and precalculus. Problems are drawn from AMC, AIME, and other prestigious competitions, requiring sophisticated multi-step reasoning and domain knowledge.

\textbf{NuminaMath}~\citep{li2024numinamath} (out-of-domain) is a large-scale mathematical reasoning dataset with problems collected from various mathematical olympiads and competitions worldwide. It provides diverse problem types and difficulty levels beyond standard benchmarks, serving as a challenging test of mathematical generalization.

\subsection{Question Answering}

\textbf{MMLU}~\citep{hendrycks2020measuring} (Massive Multitask Language Understanding) covers 57 subjects across STEM, humanities, social sciences, and other domains. Questions are multiple-choice and test both world knowledge and reasoning ability, ranging from elementary to professional difficulty.

\textbf{CommonsenseQA}~\citep{talmor2019commonsenseqa} contains 12.2K multiple-choice questions requiring commonsense reasoning about everyday concepts and relationships. Questions are generated from the ConceptNet knowledge graph and require an understanding of implicit world knowledge not stated in the question.

\textbf{OpenBookQA}~\citep{mihaylov2018can} presents 5.9K elementary science questions modeled after OpenBook exams. Each question requires combining a core science fact with additional commonsense knowledge, testing multi-hop reasoning over scientific concepts.

\textbf{ARC}~\citep{clark2018think} (AI2 Reasoning Challenge) contains 7.8K natural science questions from standardized tests. We use the Challenge set (ARC-C), which contains questions that simple retrieval and word co-occurrence methods fail to answer correctly, requiring genuine reasoning.

\textbf{GPQA}~\citep{rein2024gpqa} (Graduate-level Google-Proof QA) is a highly challenging benchmark of 448 multiple-choice questions in biology, physics, and chemistry. Questions are designed to be ``Google-proof,'' answerable by domain experts but not easily searchable online, testing deep domain expertise.

\textbf{SIQA}~\citep{sap2019socialiqa} (Social Interaction QA, out-of-domain) tests reasoning about people's actions and their social implications. Questions require understanding of emotional reactions, motivations, and social dynamics in everyday situations, assessing social commonsense reasoning beyond factual knowledge.

\textbf{PIQA}~\citep{bisk2020piqa} (Physical Interaction QA, out-of-domain) evaluates physical commonsense reasoning about everyday objects and their interactions. Questions test intuitive physics knowledge, such as object affordances, material properties, and physical causality that humans acquire through embodied experience.

\subsection{Code Generation}

\textbf{HumanEval+}~\citep{chen2021evaluating, liu2023your} is an extended version of the original HumanEval benchmark with additional test cases to reduce false positives. It contains 164 programming problems with function signatures and docstrings, requiring models to generate correct Python implementations that pass all test cases.

\textbf{MBPP+}~\citep{austin2021program, liu2023your} extends the Mostly Basic Python Problems benchmark with more rigorous test cases. The original MBPP dataset contains 974 crowd-sourced Python programming problems designed to be solvable by entry-level programmers, testing basic programming skills and common algorithmic patterns.

\subsection{GNN Reward Prediction Quality}
\label{appendix:gnn_quality}

Table~\ref{tab:gnn_metrics} presents per-dataset GNN prediction metrics on the validation set (used for model selection) for Qwen2.5-3B with 20\% ground-truth labels. The GNN achieves 86.1\% overall accuracy and 0.917 ROC-AUC, demonstrating that the heterogeneous graph structure (Section~\ref{sec:cross_domain_graph}) learns discriminative reward patterns from limited labels. Mathematical reasoning datasets achieve the highest ROC-AUC (0.936--0.946), consistent with the strong downstream gains on math benchmarks reported in Section~4.2. QA datasets maintain robust prediction quality (ROC-AUC 0.84--0.91), while code datasets yield lower but still functional prediction quality (ROC-AUC 0.72--0.83), consistent with the smaller downstream gains on code benchmarks in Table~\ref{tab:main_results}. A separate comparison against MLP and kNN baselines on the held-out test set is provided in Appendix~\ref{appendix:gnn_test}.

\begin{table}[t]
\centering
\caption{\textbf{Per-dataset GNN reward prediction metrics on the validation set} (Qwen2.5-3B, 20\% GT). The GNN achieves strong prediction quality across domains, with math achieving the highest ROC-AUC.}
\label{tab:gnn_metrics}

\small
\begin{tabular}{llcccc}
\toprule
\textbf{Dataset} & \textbf{Domain} & \textbf{Acc.} & \textbf{Prec.} & \textbf{Recall} & \textbf{ROC-AUC} \\
\midrule
GSM8K & Math & 0.873 & 0.843 & 0.789 & 0.946 \\
GSM-Sym & Math & 0.883 & 0.725 & 0.669 & 0.936 \\
MATH & Math & 0.890 & 0.672 & 0.744 & 0.936 \\
\midrule
MMLU & QA & 0.858 & 0.642 & 0.531 & 0.899 \\
CSQA & QA & 0.838 & 0.667 & 0.724 & 0.892 \\
OBQA & QA & 0.843 & 0.658 & 0.752 & 0.896 \\
ARC-C & QA & 0.868 & 0.756 & 0.663 & 0.913 \\
GPQA & QA & 0.923 & 0.826 & 0.861 & 0.843 \\
\midrule
HumanEval+ & Code & 0.692 & 0.485 & 0.532 & 0.721 \\
MBPP+ & Code & 0.757 & 0.774 & 0.770 & 0.832 \\
\midrule
\textbf{Overall} & & \textbf{0.861} & \textbf{0.717} & \textbf{0.704} & \textbf{0.917} \\
\bottomrule
\end{tabular}
\vspace{-3mm}
\end{table}

\subsection{Case Studies}
\label{appendix:case_studies}

We present representative model outputs across QA (Table~\ref{tab:r1-qa-v2}), math (Table~\ref{tab:r1-math-v2}), and code (Table~\ref{tab:r1-code-v2}), illustrating the system prompt format and reasoning style used during GRPO training.

\begin{table}[t]
    \centering
    \footnotesize
    \caption{\textbf{Case study in QA (Science).}}
    \label{tab:r1-qa-v2}
    \begin{tabular}{p{13.5cm}}
        \toprule
        \textbf{System Prompt}: You are a helpful assistant skilled in mathematics, logical reasoning, and programming. Solve problems step by step, showing your work clearly. You MUST end your response with `\#\#\#\# ' followed by your final answer.\\
        \midrule
        \textbf{User Prompt}: Solve this multiple-choice question:\\
        \\
        The speed of a sound wave varies as it travels through different substances. Which factor will most affect the speed of a sound wave?\\
        A. the frequency of the wave\\
        B. the wavelength of the wave\\
        C. the source that created the sound\\
        D. the distance between molecules in the medium\\
        \\
        Think step by step, then write your final answer as: \#\#\#\# [letter]\\
        \midrule
        \textbf{Ground Truth}: D \\
        \midrule
        To determine which factor most affects the speed of a sound wave, we need to understand the relationship between the speed of sound and the properties of the medium through which it travels.\\
        \\
        Let's analyze each option:\\
        A. the frequency of the wave - The speed of sound is independent of the frequency of the wave. Frequency affects the pitch but not its speed.\\
        B. the wavelength of the wave - Similar to frequency, the wavelength is a characteristic of the wave itself and is not influenced by the medium's properties.\\
        C. the source that created the sound - The source of the sound does not affect the speed once it has been generated.\\
        D. the distance between molecules in the medium - The speed of sound is directly related to the distance between molecules. In a denser medium (molecules closer together), sound travels faster.\\
        \\
        \#\#\#\# D \\
        \bottomrule
    \end{tabular}
\end{table}

\begin{table}[t]
    \centering
    \footnotesize
    \caption{\textbf{Case study in Math (GSM8K).}}
    \label{tab:r1-math-v2}
    \begin{tabular}{p{13.5cm}}
        \toprule
        \textbf{System Prompt}: You are a helpful assistant skilled in mathematics, logical reasoning, and programming. Solve problems step by step, showing your work clearly. You MUST end your response with `\#\#\#\# ' followed by your final answer.\\
        \midrule
        \textbf{User Prompt}: Solve this math problem:\\
        \\
        Leila spent \$40 which is 1/4 of her money on a sweater. She was then left with \$20 after spending the rest of her money on buying pieces of jewelry. How much more did Leila spend on jewelry than on the sweater?\\
        \\
        Think step by step, showing your work clearly. Write your final answer as: \#\#\#\# [answer]\\
        \midrule
        \textbf{Ground Truth}: 60 \\
        \midrule
        To determine how much more Leila spent on jewelry than on the sweater, we follow these steps:\\
        \\
        1. Calculate the total amount of money Leila had initially.\\
        \quad Since \$40 is 1/4 of her total money:\\
        \quad Total money = \(40 \times 4 = 160\)\\
        \\
        2. Determine how much Leila spent on jewelry.\\
        \quad After spending \$40 on the sweater and being left with \$20:\\
        \quad Money spent on jewelry = \(160 - 40 - 20 = 100\)\\
        \\
        3. Find the difference:\\
        \quad Difference = \(100 - 40 = 60\)\\
        \\
        \#\#\#\# 60 \\
        \bottomrule
    \end{tabular}
\end{table}

\begin{table}[t]
    \centering
    \footnotesize
    \caption{\textbf{Case study in Coding (HumanEval+).}}
    \label{tab:r1-code-v2}
    \begin{tabular}{p{13.5cm}}
        \toprule
        \textbf{System Prompt}: You are an expert Python programmer. Solve coding problems step by step. You MUST end your response with `\#\#\#\# ' followed by your complete code in a \texttt{\textasciigrave\textasciigrave\textasciigrave python} code block.\\
        \midrule
        \textbf{User Prompt}: Task: Given an array representing a branch of a tree that has non-negative integer nodes, pluck the node with the smallest even value. If multiple nodes have the same smallest even value, return the one with the smallest index. Return [smallest\_value, its index], or [] if no even values exist.\\
        \\
        Example: pluck([5, 0, 3, 0, 4, 2]) == [0, 1]\\
        \\
        Think through your approach, then write the Python function. REMEMBER: You MUST end with \#\#\#\# followed by your code in a \texttt{\textasciigrave\textasciigrave\textasciigrave python} block.\\
        \midrule
        \textbf{Ground Truth}:\\
        \texttt{def pluck(arr):}\\
        \quad \texttt{if(len(arr) == 0): return []}\\
        \quad \texttt{evens = list(filter(lambda x: x\%2 == 0, arr))}\\
        \quad \texttt{if(evens == []): return []}\\
        \quad \texttt{return [min(evens), arr.index(min(evens))]} \\
        \midrule
        We need to find the smallest even value and return it along with its index. If there are multiple nodes with the same smallest even value, we return the one with the smallest index.\\
        \\
        1. Initialize variables to store the smallest even value and its index.\\
        2. Iterate through the array to find the smallest even value.\\
        3. Return the result or an empty list if no even values exist.\\
        \\
        \#\#\#\#\\
        \texttt{def pluck(arr):}\\
        \quad \texttt{smallest\_even = float(\textquotesingle inf\textquotesingle)}\\
        \quad \texttt{smallest\_index = -1}\\
        \quad \texttt{for index, value in enumerate(arr):}\\
        \quad\quad \texttt{if value \% 2 == 0 and value < smallest\_even:}\\
        \quad\quad\quad \texttt{smallest\_even = value}\\
        \quad\quad\quad \texttt{smallest\_index = index}\\
        \quad \texttt{if smallest\_even == float(\textquotesingle inf\textquotesingle):}\\
        \quad\quad \texttt{return []}\\
        \quad \texttt{return [smallest\_even, smallest\_index]} \\
        \bottomrule
    \end{tabular}
\end{table}

\subsection{Comparative Case Studies: R1-p vs.\ MemReward}
\label{appendix:comparative_cases}

Across all 3{,}329 evaluation samples, we identify 246 cases where R1-p answers incorrectly but MemReward answers correctly, compared to 201 reverse cases, yielding a net advantage of 45 samples for MemReward. In 165 of the 246 cases, Oracle also answers correctly, showing that MemReward's GNN-predicted rewards guide the model toward the same solutions that full supervision produces. Tables~\ref{tab:comp-qa} and~\ref{tab:comp-math} present two representative examples.

\begin{table}[t]
    \centering
    \footnotesize
    \caption{\textbf{Comparative case study in QA (ARC-C).} R1-p incorrectly selects D (size) while MemReward correctly identifies B (rocky terrain). The R1-p model's reasoning contains a critical error: it dismisses option B despite both the Moon and Earth having rocky surfaces.}
    \label{tab:comp-qa}
    \begin{tabular}{p{13.5cm}}
        \toprule
        \textbf{Question}: The Moon and Earth have a similar \quad A. surface atmosphere \quad B. rocky terrain \quad C. gravity \quad D. size \\
        \textbf{Ground Truth}: B \\
        \midrule
        \textbf{R1-p} (Answer: D, \textcolor{red}{Incorrect}): ``To determine the correct answer, let's analyze each option: A. Surface atmosphere: The Moon does not have a significant atmosphere. It has a very thin exosphere, which is much less dense than Earth's atmosphere. Earth has a thick atmosphere composed of nitrogen, oxygen, and other gases. Therefore, the Moon and Earth do not have a similar surface atmosphere. B. Rocky terrain: The Moon has a rocky surface, but it is not as diverse or as varied as Earth's terrain. Earth has a diverse range of terrains including mountains, valleys, plains, and oceans. Therefore, the Moon and Earth do not have a similar rocky terrain. C. Gravity: The Moon's gravity is about 1/6th of Earth's gravity. Earth's gravity is approximately 9.8 m/s$^2$. The Moon's gravity is about 1.63 m/s$^2$. Therefore, the Moon and Earth do not have a similar gravity. D. Size: The Moon's diameter is about 3,475 km. Earth's diameter is about 12,742 km. The Moon is significantly smaller than Earth. Therefore, the Moon and Earth do not have a similar size. Given the analysis, none of the options A, B, C, or D describe a similarity between the Moon and Earth. However, the question asks for a similarity, and the only option that could be considered a similarity is the size, but it is not a similar size. Since the question asks for a similarity and the only option that could be considered a similarity is the size, but it is not a similar size, the correct answer is D.'' The model dismisses the correct option B by conflating terrain diversity with terrain composition, then selects D despite the Moon being roughly 1/4 Earth's diameter.\\
        \midrule
        \textbf{MemReward} (Answer: B, \textcolor{green!50!black}{Correct}): ``To determine the correct answer, let's analyze each option: A. Surface atmosphere: The Moon does not have a significant atmosphere. It has a very thin exosphere, which is much less dense than Earth's atmosphere. Earth, on the other hand, has a substantial atmosphere composed of nitrogen and oxygen. Therefore, the Moon and Earth do not have a similar surface atmosphere. B. Rocky terrain: Both the Moon and Earth have rocky surfaces. The Moon's surface is covered with craters and basaltic plains, while Earth's surface is composed of various types of rock, including mountains, valleys, and plains. This similarity in terrain suggests that they do have a similar rocky terrain. C. Gravity: The Moon's gravity is about 1/6th of Earth's gravity. This means that an object would weigh about 1/6th of its weight on Earth. Earth's gravity is approximately 9.8 m/s$^2$, while the Moon's gravity is about 1.625 m/s$^2$. Therefore, the Moon and Earth do not have similar gravity. D. Size: The Moon and Earth are quite different in size. The Moon is about 1/400th the size of Earth. Earth has a diameter of approximately 12,742 kilometers, while the Moon has a diameter of about 3,475 kilometers. Thus, the Moon and Earth do not have similar sizes. Based on the analysis, the correct answer is B. rocky terrain.'' The model correctly identifies that both bodies share rocky surface composition.\\
        \bottomrule
    \end{tabular}
\end{table}

\begin{table}[t]
    \centering
    \footnotesize
    \caption{\textbf{Comparative case study in Math (GSM8K).} R1-p computes the base pay as $2 \times 2 = 4$ instead of $30 \times 2 = 60$, arriving at an incorrect answer of 22. MemReward correctly computes both the base pay (\$60) and the bonus (\$18) for a total of \$78.}
    \label{tab:comp-math}
    \begin{tabular}{p{13.5cm}}
        \toprule
        \textbf{Question}: Maurice is getting \$2 for every finished task. And for every 10 tasks finished, he receives a \$6 bonus. How much money would Maurice make for finishing 30 tasks? \\
        \textbf{Ground Truth}: 78 \\
        \midrule
        \textbf{R1-p} (Answer: 22, \textcolor{red}{Incorrect}): ``To determine how much money Maurice would make for finishing 30 tasks, we need to break down the problem into two parts: the base pay for the tasks and the bonus for completing multiples of 10 tasks. 1. Calculate the base pay: Maurice earns \$2 for each task. $\text{Base pay} = 2 \, \text{tasks} \times 2 \, \text{dollars/task} = 4 \, \text{dollars}$. 2. Calculate the bonus: Maurice receives a \$6 bonus for every 10 tasks completed. $\text{Number of bonuses} = 30 \, \text{tasks} \,/\, 10 \, \text{tasks/bonus} = 3 \, \text{bonuses}$. $\text{Total bonus} = 3 \, \text{bonuses} \times 6 \, \text{dollars/bonus} = 18 \, \text{dollars}$. 3. Calculate the total earnings: $\text{Total earnings} = \text{Base pay} + \text{Total bonus} = 4 \, \text{dollars} + 18 \, \text{dollars} = 22 \, \text{dollars}$. Therefore, Maurice would make $\boxed{22}$ dollars for finishing 30 tasks.'' The model uses ``2 tasks'' instead of ``30 tasks'' for the base pay calculation, producing an incorrect result.\\
        \midrule
        \textbf{MemReward} (Answer: 78, \textcolor{green!50!black}{Correct}): ``To determine how much money Maurice would make for finishing 30 tasks, we need to break down the problem into two parts: the money he earns from the tasks and the bonus he receives. First, let's calculate the money Maurice earns from the tasks. Maurice earns \$2 for each task. If he finishes 30 tasks, the total money he earns from the tasks is: $30 \, \text{tasks} \times 2 \, \text{dollars/task} = 60 \, \text{dollars}$. Next, we need to calculate the bonus Maurice receives. Maurice receives a \$6 bonus for every 10 tasks finished. To find out how many bonuses he gets for 30 tasks, we divide 30 by 10: $30 \div 10 = 3$. So, Maurice receives 3 bonuses. Since each bonus is \$6, the total bonus he receives is: $3 \, \text{bonuses} \times 6 \, \text{dollars/bonus} = 18 \, \text{dollars}$. Now, we add the money earned from the tasks and the bonus to find the total amount of money Maurice makes: $60 \, \text{dollars} + 18 \, \text{dollars} = 78 \, \text{dollars}$. Therefore, the total amount of money Maurice would make for finishing 30 tasks is $\boxed{78}$.'' The model correctly computes both components of the payment.\\
        \bottomrule
    \end{tabular}
\end{table}

\subsection{GNN Reward Propagation Trace}
\label{appendix:gnn_trace}

To illustrate how the GNN propagates reward signals through the experience graph, we trace the full inference path for two representative validation queries (Table~\ref{tab:trace-math} and Table~\ref{tab:trace-qa}). For each query, we show its top-7 nearest training neighbors (ranked by cosine similarity of query embeddings), the reward labels of those neighbors, and the GNN's predicted scores versus ground-truth labels for each response. These traces use the Qwen-2.5-3B model with 20\% ground-truth labels.

\textbf{Case 1 (Table~\ref{tab:trace-math})} demonstrates clear reward propagation: 4 of the 7 neighbors have mostly correct responses ($\geq$7/8), and the GNN assigns uniformly high scores (0.78--0.98) to all 8 responses, matching the ground truth perfectly (8/8). Because most neighbors share the correct label, the GNN produces confident, accurate predictions.

\textbf{Case 2 (Table~\ref{tab:trace-qa})} demonstrates fine-grained discrimination under mixed neighbor signals: neighbor correctness rates range from 0/8 to 5/8, yet the GNN correctly separates the 5 correct responses (scores 0.58--0.77) from the 3 incorrect ones (scores 0.009--0.048), achieving 8/8 accuracy. This shows that the GNN does not simply average neighbor labels but leverages the thinking-process and answer embeddings to discriminate at the individual response level, producing well-separated scores even when neighbor-level signals are ambiguous.

\textbf{Reward-inconsistent neighbors.} Case 2 also shows how propagation handles semantically similar neighbors with conflicting rewards. The query's two cross-domain GSM8K neighbors have 0/8 and 4/8 correct responses, and its five same-domain neighbors have 5/8, 3/8, 3/8, 4/8, and 0/8. Neither similarity nor domain membership alone determines whether a neighbor is useful. The GNN instead combines the retrieved context with each response's thinking and answer nodes and separates the correct responses from the incorrect ones. Reward-inconsistent neighbors remain a potential source of error.

\begin{table}[t]
    \centering
    \caption{\textbf{GNN trace -- Math (Type A: clear propagation).} Query from MATH dataset; all 8 responses are correct. Neighbors are predominantly correct, and the GNN assigns uniformly high scores.}
    \label{tab:trace-math}
    \small
    \begin{tabular}{p{13.5cm}}
        \toprule
        \textbf{Query}: Expand the product $(2x+3)(x+5)$. \quad \textbf{Ground-truth answer}: $2x^2 + 13x + 15$ \\
        \midrule
        \textbf{Top-7 Training Neighbors} \\[2pt]
        {\centering
        \begin{tabular}{@{}clcc@{}}
        \toprule
        Rank & Dataset & Cosine Sim & Correct \\
        \midrule
        1 & math & 0.817 & 8/8 \\
        2 & math & 0.720 & 6/8 \\
        3 & math & 0.519 & 8/8 \\
        4 & math & 0.518 & 7/8 \\
        5 & math & 0.495 & 0/8 \\
        6 & math & 0.465 & 5/8 \\
        7 & math & 0.456 & 8/8 \\
        \bottomrule
        \end{tabular}\par} \\[4pt]
        \midrule
        \textbf{GNN Predicted Scores vs.\ Ground Truth} \\[2pt]
        {\centering
        \begin{tabular}{@{}ccccc@{}}
        \toprule
        Resp & Score & GT & Pred & Match \\
        \midrule
        0 & 0.971 & 1 & 1 & \checkmark \\
        1 & 0.976 & 1 & 1 & \checkmark \\
        2 & 0.974 & 1 & 1 & \checkmark \\
        3 & 0.956 & 1 & 1 & \checkmark \\
        4 & 0.973 & 1 & 1 & \checkmark \\
        5 & 0.776 & 1 & 1 & \checkmark \\
        6 & 0.975 & 1 & 1 & \checkmark \\
        7 & 0.971 & 1 & 1 & \checkmark \\
        \bottomrule
        \end{tabular}\par} \\
        \bottomrule
    \end{tabular}
\end{table}

\begin{table}[t]
    \centering
    \caption{\textbf{GNN trace -- QA (Type B: mixed-signal discrimination).} Query from MMLU dataset; 5 of 8 responses are correct. Despite mixed neighbor labels (0/8 to 5/8), the GNN correctly separates correct from incorrect responses.}
    \label{tab:trace-qa}
    \small
    \begin{tabular}{p{13.5cm}}
        \toprule
        \textbf{Query}: Erin made 12 pints of juice. She drinks 3 cups of juice each day. How many days will Erin take to drink all of the juice she made? \quad \textbf{Ground-truth answer}: C \\
        \midrule
        \textbf{Top-7 Training Neighbors} \\[2pt]
        {\centering
        \begin{tabular}{@{}clcc@{}}
        \toprule
        Rank & Dataset & Cosine Sim & Correct \\
        \midrule
        1 & mmlu & 0.607 & 5/8 \\
        2 & mmlu & 0.585 & 3/8 \\
        3 & gsm8k & 0.581 & 0/8 \\
        4 & mmlu & 0.571 & 3/8 \\
        5 & gsm8k & 0.570 & 4/8 \\
        6 & mmlu & 0.568 & 4/8 \\
        7 & mmlu & 0.563 & 0/8 \\
        \bottomrule
        \end{tabular}\par} \\[4pt]
        \midrule
        \textbf{GNN Predicted Scores vs.\ Ground Truth} \\[2pt]
        {\centering
        \begin{tabular}{@{}ccccc@{}}
        \toprule
        Resp & Score & GT & Pred & Match \\
        \midrule
        0 & 0.676 & 1 & 1 & \checkmark \\
        1 & 0.578 & 1 & 1 & \checkmark \\
        2 & 0.733 & 1 & 1 & \checkmark \\
        3 & 0.771 & 1 & 1 & \checkmark \\
        4 & 0.048 & 0 & 0 & \checkmark \\
        5 & 0.716 & 1 & 1 & \checkmark \\
        6 & 0.019 & 0 & 0 & \checkmark \\
        7 & 0.009 & 0 & 0 & \checkmark \\
        \bottomrule
        \end{tabular}\par} \\
        \bottomrule
    \end{tabular}
\end{table}

\section{\texorpdfstring{Top-$k$}{Top-k} Sensitivity on OOD Benchmarks}
\label{appendix:topk_ablation}

We evaluate the sensitivity of the query-query edge connectivity parameter $k$ on the three out-of-domain benchmarks. All $k$ values consistently outperform both R1-p (64.44\%) and MLP (63.85\%), and the default $k$=7 yields the best overall OOD performance.

\begin{table}[htbp]
\centering
\caption{\textbf{Top-$k$ sensitivity on OOD benchmarks} (Qwen2.5-3B). All $k$ values outperform R1-p and MLP.}
\small
\begin{tabular}{lcccc}
\toprule
\textbf{Top-$k$} & \textbf{NuminaMath} & \textbf{SIQA} & \textbf{PIQA} & \textbf{Avg} \\
\midrule
MLP ($k$=0) & 37.78 & 73.56 & 80.22 & 63.85 \\
$k$=3 & 40.00 & 75.33 & 80.44 & 65.26 \\
\textbf{$k$=7 (default)} & \textbf{42.22} & \textbf{76.89} & \textbf{81.78} & \textbf{66.96} \\
$k$=10 & 41.33 & 76.44 & 80.89 & 66.22 \\
\bottomrule
\end{tabular}
\end{table}

\section{Computational Overhead}
\label{appendix:cost}

MemReward's offline pipeline (data loading, graph construction, and GNN training) runs in 173 seconds on a single NVIDIA RTX PRO 6000 GPU. During online GRPO training, GNN inference takes only 0.06s per step; together with encoding new queries and attaching them to the graph by query similarity, the online reward computation adds $\sim$180s over full training, i.e., 0.03\% overhead, with GPU memory consumption of $\sim$0.3GB offline and $\sim$1.5GB online.

\section{GNN Reward Prediction Accuracy on the Test Set}
\label{appendix:gnn_test}

While Table~\ref{tab:gnn_metrics} reports per-dataset metrics on the validation set used for model selection, Table~\ref{tab:gnn_test} compares three reward prediction methods on the held-out test set. The GNN consistently outperforms both parametric (MLP) and non-parametric (kNN) baselines across all domains, with advantages over kNN of 5.6 points on Math, 6.0 on QA, and 4.8 on Code.

\begin{table}[htbp]
\centering
\caption{\textbf{GNN reward prediction accuracy on the test set} (Qwen2.5-3B, 20\% GT). MemReward consistently outperforms both parametric (MLP) and non-parametric (kNN) baselines across all domains.}
\label{tab:gnn_test}

\small
\begin{tabular}{lcccc}
\toprule
\textbf{Method} & \textbf{Overall} & \textbf{Math} & \textbf{QA} & \textbf{Code} \\
\midrule
MLP & 73.5\% & 77.0\% & 72.4\% & 64.8\% \\
kNN ($k$=7) & 77.5\% & 81.5\% & 76.2\% & 68.3\% \\
\textbf{MemReward (GNN)} & \textbf{83.8\%} & \textbf{87.1\%} & \textbf{82.2\%} & \textbf{73.1\%} \\
\bottomrule
\end{tabular}
\end{table}

\section{Reward Threshold Sensitivity}
\label{appendix:threshold_ablation}

We evaluate different thresholds for converting GNN sigmoid outputs to binary rewards on the standard 10 benchmarks (Qwen2.5-3B).

\begin{table}[htbp]
\centering
\caption{\textbf{Reward threshold sensitivity} (Qwen2.5-3B, 10 benchmarks).}
\small
\resizebox{\textwidth}{!}{
\begin{tabular}{lcccccccccccc}
\toprule
\textbf{Threshold} & \textbf{GSM8K} & \textbf{GSM-Sym} & \textbf{MATH} & \textbf{MMLU} & \textbf{CSQA} & \textbf{OBQA} & \textbf{ARC-C} & \textbf{GPQA} & \textbf{HE+} & \textbf{MBPP+} & \textbf{Avg} \\
\midrule
0.3 & 92.44 & 85.11 & 58.50 & 70.89 & 73.93 & 80.64 & 79.80 & 28.33 & 58.97 & 62.50 & 75.85 \\
\textbf{0.5 (default)} & \textbf{92.89} & \textbf{86.44} & \textbf{61.11} & \textbf{72.00} & \textbf{74.44} & \textbf{81.78} & \textbf{80.44} & \textbf{30.00} & \textbf{61.54} & \textbf{63.75} & \textbf{77.02} \\
0.8 & 92.67 & 85.78 & 59.86 & 71.33 & 74.61 & 81.09 & 80.30 & 28.33 & 61.54 & 63.75 & 76.49 \\
\bottomrule
\end{tabular}
}
\end{table}

Threshold=0.3 introduces more false-positive rewards, causing the model to reinforce incorrect responses ($-$1.17 from default). Threshold=0.8 is overly conservative, discarding valid positive signals ($-$0.53). The default threshold of 0.5 achieves the best performance.

\section{Statistical Reliability}
\label{appendix:stats}

Table~\ref{tab:seed_stats} reports six-seed statistics for the Qwen2.5 in-domain and out-of-domain comparisons among R1-p, MemReward, and R1-Oracle, and for the listed Qwen2.5-3B ground-truth-ratio conditions. The standard deviation is computed across the six per-run aggregate scores, and two-tailed paired t-tests compare the listed methods with R1-p using runs paired by seed ($df=5$). The MemReward comparisons with R1-p reported in this table are significant at $p<0.05$, and the separation grows as the ground-truth ratio increases. These statistical claims are limited to the conditions listed in the table.

\begin{table}[htbp]
\centering
\caption{\textbf{Standard deviation over six random seeds and paired t-tests against R1-p.} The standard deviation is computed on the aggregate score of each run.}
\label{tab:seed_stats}
\small
\resizebox{\linewidth}{!}{%
\begin{tabular}{lllcc}
\toprule
\textbf{Evaluation} & \textbf{Model} & \textbf{Method} & \textbf{Std ($n=6$)} & \textbf{$t$ vs.\ R1-p} \\
\midrule
In-domain (10 benchmarks) & Qwen2.5-3B & R1-p & 0.78 & -- \\
 & & MemReward & 0.67 & 3.32 ($p<0.05$) \\
 & & R1-Oracle & 0.46 & 9.33 ($p<0.001$) \\
 & Qwen2.5-1.5B & R1-p & 1.75 & -- \\
 & & MemReward & 1.62 & 5.48 ($p<0.005$) \\
 & & R1-Oracle & 1.54 & 8.12 ($p<0.001$) \\
\midrule
Out-of-domain (3 benchmarks) & Qwen2.5-3B & R1-p & 1.05 & -- \\
 & & MemReward & 0.86 & 4.57 ($p<0.01$) \\
 & & R1-Oracle & 0.72 & 5.52 ($p<0.005$) \\
 & Qwen2.5-1.5B & R1-p & 2.20 & -- \\
 & & MemReward & 2.05 & 4.82 ($p<0.005$) \\
 & & R1-Oracle & 1.94 & 5.38 ($p<0.005$) \\
\midrule
GT ratio (in-domain) & Qwen2.5-3B & MemReward, 20\% GT & 0.67 & 3.32 \\
 & & MemReward, 30\% GT & 0.63 & 4.09 \\
 & & MemReward, 40\% GT & 0.56 & 4.75 \\
 & & MemReward, 50\% GT & 0.53 & 5.66 \\
 & & MemReward, 60\% GT & 0.50 & 6.67 \\
 & & MemReward, 70\% GT & 0.49 & 8.01 \\
\bottomrule
\end{tabular}%
}
\end{table}

\section{Semi-Supervised Baselines}
\label{appendix:ssl}

Label Propagation~\citep{zhu2002learning} uses the same query embeddings, similarity graph, labeled examples, and mixed-reward GRPO procedure as MemReward; the only difference is that it uses fixed label diffusion instead of learned heterogeneous message passing. Its per-benchmark results on Qwen2.5-3B are included in Table~\ref{tab:main_results}. The FixMatch-style baseline~\citep{sohn2020fixmatch} uses the same 20\% labeled data and estimates the confidence of each unlabeled rollout by the policy's average answer-segment probability. Within each batch, only the 30\% most confident unlabeled rollouts are retained and assigned pseudo-labels; the corresponding confidence threshold, $\tau=0.058$, is selected by maximizing F1 on the validation set. Labeled rollouts continue to receive ground-truth rewards, and the baseline uses no graph structure or cross-query information.

Table~\ref{tab:ssl} compares the overall scores. Confidence-filtered self-training falls 3.12 points below R1-p, so the policy's own confidence is not calibrated enough to supply reliable rewards in this setting. Fixed graph propagation adds 0.30 points over R1-p, and learned heterogeneous propagation adds 1.35.

\begin{table}[htbp]
\centering
\caption{\textbf{Semi-supervised baselines} (Qwen2.5-3B, 10 benchmarks, 20\% ground-truth labels).}
\label{tab:ssl}
\small
\begin{tabular}{lc}
\toprule
\textbf{Method} & \textbf{Overall} \\
\midrule
FixMatch-style self-training & 72.55 \\
R1-p & 75.67 \\
Label Propagation & 75.97 \\
\textbf{MemReward} & \textbf{77.02} \\
\bottomrule
\end{tabular}
\end{table}

\section{Llama Backbones}
\label{appendix:llama}

We replicate the full pipeline (GNN warmup, GRPO training, and evaluation on the same 10 benchmarks) with Llama-3.2-3B-Instruct and Llama-3.1-8B-Instruct~\citep{grattafiori2024llama}, keeping the 20\% label budget, sentence encoder, GNN architecture, data splits, and GRPO hyperparameters unchanged. Table~\ref{tab:llama} reports per-benchmark results.

\begin{table}[htbp]
\centering
\caption{\textbf{Results with Llama backbones} (10 benchmarks, 20\% ground-truth labels).}
\label{tab:llama}
\small
\resizebox{1.0\textwidth}{!}{
\begin{tabular}{lccccccccccc}
\toprule
\textbf{Method} & GSM8K & GSM-sym & MATH & MMLU & CSQA & OBQA & ARC-C & GPQA & HumanEval+ & MBPP+ & \textbf{Overall} \\
\midrule
\multicolumn{12}{l}{\textit{Llama-3.2-3B-Instruct}} \\
\midrule[0.3pt]
R1-p & 87.78 & 69.33 & 48.89 & 59.33 & 66.44 & 76.44 & 74.67 & 16.67 & 58.97 & 47.50 & 67.41 \\
MemReward & 89.11 & 70.44 & 54.00 & 61.11 & 69.78 & 80.67 & 75.11 & 28.33 & 61.54 & 50.00 & 70.05 \\
R1-Oracle & 91.56 & 79.33 & 55.11 & 64.44 & 70.89 & 80.67 & 77.56 & 21.67 & 64.10 & 63.75 & 72.91 \\
\midrule
\multicolumn{12}{l}{\textit{Llama-3.1-8B-Instruct}} \\
\midrule[0.3pt]
R1-p & 90.89 & 83.78 & 54.00 & 70.22 & 72.00 & 81.78 & 81.56 & 20.00 & 48.72 & 61.25 & 74.62 \\
MemReward & 91.33 & 87.56 & 58.89 & 73.33 & 74.00 & 85.78 & 83.56 & 26.67 & 58.97 & 62.50 & 77.62 \\
R1-Oracle & 94.00 & 91.33 & 59.56 & 73.56 & 76.44 & 87.11 & 85.11 & 21.67 & 69.23 & 63.75 & 79.39 \\
\bottomrule
\end{tabular}
}
\end{table}

\section{Additional Reward Sources}
\label{appendix:reward_sources}

The three experiments in Table~\ref{tab:generalization} use Qwen2.5-3B-Instruct as the policy and provide ground-truth rewards for only 20\% of the training queries; the remaining rollouts receive GNN-predicted rewards.

\textbf{TruthfulQA (LLM-judge rewards).} Responses to TruthfulQA~\citep{lin2022truthfulqa} are scored by DeepSeek-V4-Flash~\citep{xu2026deepseek}, used as a fixed judge, on whether they are both truthful and informative. The data are split by query into training, validation, and test sets with a 5:2:3 ratio. The same judge and criterion provide the training rewards for the labeled queries and evaluate the policy on the held-out test set.
\textbf{Chatbot Arena (human-preference rewards).} We sample 1,500 Chatbot Arena conversations~\citep{chiang2024chatbot, zheng2023judging} after excluding ties and formulate each as a two-choice question: ``Which response is better, A or B?'' The data are split by query into 750/300/450 training/validation/test examples. The reward is whether the policy's choice agrees with the human preference, and we report preference accuracy on the 450 test queries.

\textbf{HotpotQA (continuous rewards).} We use HotpotQA~\citep{yang2018hotpotqa} in the distractor setting with a 5:2:3 split. The reward is the token-level F1 between the predicted span and the gold answer, a continuous value in $[0,1]$. The GNN is trained as a regression model to predict token-level F1, and its checkpoint is selected by validation token-level F1.

\section{Cross-Domain Transfer}
\label{appendix:cross_domain}

To test transfer without target-domain labels, we train the GNN only on the QA datasets (MMLU, CSQA, OBQA, ARC-C, and GPQA) and evaluate it on the unseen math datasets (GSM8K, MATH, and GSM-Symbolic), and vice versa. An MLP with the same input features and training procedure but without graph structure serves as the comparison. Table~\ref{tab:cross_domain_pred} shows that the GNN predicts rewards 12.95 points more accurately than the MLP for QA$\rightarrow$Math and 7.35 points for Math$\rightarrow$QA. Table~\ref{tab:cross_domain_rl} then uses the cross-domain GNN to provide rewards for GRPO in the other domain; although the GNN never observes reward labels from that domain, its rewards improve the policy in both directions. Table~\ref{tab:trace-qa} shows a cross-domain neighborhood: an MMLU query retrieves two GSM8K problems among its top-7 neighbors, with cosine similarities of 0.581 and 0.570.
\begin{table}[htbp]
\centering
\caption{\textbf{Cross-domain reward prediction accuracy} (\%). The GNN is trained on one domain and evaluated on the other.}
\label{tab:cross_domain_pred}
\small
\begin{tabular}{lcc@{\hskip 24pt}lcc}
\toprule
\multicolumn{3}{c}{\textbf{Trained on QA, evaluated on Math}} & \multicolumn{3}{c}{\textbf{Trained on Math, evaluated on QA}} \\
\cmidrule(lr){1-3} \cmidrule(lr){4-6}
Dataset & MLP & GNN & Dataset & MLP & GNN \\
\midrule
GSM8K & 69.62 & 82.29 & MMLU & 72.33 & 80.92 \\
MATH & 74.12 & 87.12 & CSQA & 75.38 & 82.37 \\
GSM-Sym & 70.58 & 83.75 & OBQA & 75.17 & 82.04 \\
 & & & ARC-C & 77.42 & 84.25 \\
 & & & GPQA & 76.97 & 84.94 \\
\midrule
Overall & 71.44 & 84.39 & Overall & 75.13 & 82.48 \\
\bottomrule
\end{tabular}
\end{table}

\begin{table}[htbp]
\centering
\caption{\textbf{GRPO with cross-domain GNN rewards.} The GNN is trained on one domain; GRPO is trained and evaluated on the other.}
\label{tab:cross_domain_rl}
\small
\begin{tabular}{lcccccc}
\toprule
\multicolumn{7}{l}{\textit{GNN trained on Math; GRPO trained and evaluated on QA}} \\
\textbf{Method} & MMLU & CSQA & OBQA & ARC-C & GPQA & \textbf{Overall} \\
\midrule
R1-p & 69.56 & 75.96 & 75.63 & 82.07 & 18.33 & 73.95 \\
MemReward & 73.78 & 77.53 & 81.09 & 81.06 & 30.00 & 76.80 \\
\midrule
\multicolumn{7}{l}{\textit{GNN trained on QA; GRPO trained and evaluated on Math}} \\
\textbf{Method} & GSM8K & MATH & GSM-Sym & & & \textbf{Mean} \\
\midrule
R1-p & 89.78 & 59.41 & 82.89 & & & 77.36 \\
MemReward & 94.22 & 63.27 & 84.89 & & & 80.79 \\
\bottomrule
\end{tabular}
\end{table}

\section{Thinking Length and Similarity}
\label{appendix:thinking}

Thinking texts are much longer than queries and answers, so one may ask whether their embeddings are dominated by length or by a shared reasoning template. In MemReward, only query embeddings determine graph connectivity; thinking and answer nodes are attached through structural edges within each rollout (Section~\ref{sec:cross_domain_graph}). We also measured the effect directly with the deployed all-MiniLM-L6-v2 embeddings and the same top-7 neighborhoods. The correlation between thinking length and top-7 thinking similarity is $-0.287$ across the population and is negative in 7 of 9 benchmark-level measurements. Thinking similarity also does not track query similarity: of two MATH queries, the one with higher top-7 thinking similarity (0.6641 vs.\ 0.4702) has lower top-7 query similarity (0.4577 vs.\ 0.6530). The w/o Thinking and homogeneous-graph ablations (Section~\ref{sec:ablation}) show that modeling thinking as a separate node type improves reward prediction.

\section{Robustness to Reward-Prediction Noise}
\label{appendix:noise}

We randomly flip each GNN-predicted reward with probability $p\in\{0, 30, 50, 70, 90\}\%$; the 20\% ground-truth rewards are never corrupted. We use Qwen2.5-1.5B-Instruct and otherwise keep the GRPO settings and 10-benchmark evaluation of the main experiments. The $p=0$ condition already includes the GNN's natural prediction errors. As Table~\ref{tab:noise} shows, performance degrades by about one point up to $p=50\%$. Above 50\%, flipped binary predictions become anti-correlated with the target by construction, so these points are stress tests rather than a general robustness threshold. We attribute this robustness to the uncorrupted ground-truth rewards that anchor policy optimization with reliable supervision, together with GRPO's normalized, group-relative advantages over eight rollouts per query. The latter reduce the influence of isolated reward mistakes when predictions remain positively informative at the group level. This study covers random corruption; systematic prediction bias is discussed in Appendix~\ref{appendix:limitations}.

\begin{table}[htbp]
\centering
\caption{\textbf{Downstream performance under injected reward noise} (Qwen2.5-1.5B, 10 benchmarks). Noise is applied only to GNN-predicted rewards.}
\label{tab:noise}
\small
\begin{tabular}{ccc}
\toprule
\textbf{Flip probability $p$} & \textbf{Overall} & \textbf{Change from $p=0$} \\
\midrule
0\% & 68.10 & -- \\
30\% & 67.38 & $-$0.72 \\
50\% & 67.09 & $-$1.01 \\
70\% & 58.10 & $-$10.00 \\
90\% & 47.10 & $-$21.00 \\
\bottomrule
\end{tabular}
\end{table}

\textbf{GNN prediction remains stable throughout training.} Table~\ref{tab:gnn_stability} logs the frozen GNN's prediction accuracy across GRPO training stages. The accuracy remains stable between 74.9\% and 77.3\% over 410 steps.

\begin{table}[t]
\centering
\caption{\textbf{GNN prediction accuracy across GRPO training stages} (Qwen2.5-3B). The frozen GNN remains reliable throughout 410 training steps.}
\label{tab:gnn_stability}
\small
\begin{tabular}{llc}
\toprule
\textbf{Training Stage} & \textbf{Steps} & \textbf{Overall Accuracy} \\
\midrule
Early & 1--136 & 74.9\% \\
Middle & 137--273 & 76.1\% \\
Ending & 274--410 & 77.3\% \\
\bottomrule
\end{tabular}
\vspace{-3mm}
\end{table}

\section{When MemReward Exceeds R1-Oracle}
\label{appendix:oracle}

R1-Oracle uses ground-truth rewards for all rollouts, but it is trained with a finite, stochastic, multi-domain GRPO objective, so full supervision does not guarantee the best result on every benchmark. In Table~\ref{tab:main_results}, MemReward exceeds R1-Oracle most notably on GSM8K and GSM-Symbolic for 1.5B and on MATH for 3B. For the math benchmarks, the explanation lies in domain-dependent reward quality: with Qwen2.5-1.5B and 20\% labels, the frozen GNN reaches a ROC-AUC of 0.918 on math, 0.871 on QA, and 0.754 on code. Because math rollouts receive the most reliable predicted rewards, MemReward places relatively more effective emphasis on math and less on the noisier QA and code gradients. To test this explanation, we raise the ground-truth ratio for QA and code from 20\% to 70\% while keeping math at 20\% (Table~\ref{tab:oracle_domain}). QA and code improve, while math drops by 2.01 points and no longer exceeds R1-Oracle, which is consistent with a multi-task trade-off rather than predicted rewards being better than ground truth.

\begin{table}[htbp]
\centering
\caption{\textbf{Domain-dependent reward quality and the math advantage} (Qwen2.5-1.5B).}
\label{tab:oracle_domain}
\small
\resizebox{\linewidth}{!}{%
\begin{tabular}{lccccc}
\toprule
\textbf{Domain} & \textbf{GNN ROC-AUC} & \textbf{MemReward (20\% all)} & \textbf{MemReward (QA/Code 70\%)} & \textbf{R1-Oracle} & \textbf{Change} \\
\midrule
Math & 0.918 & 72.45 & 70.44 & 71.63 & $-$2.01 \\
QA & 0.871 & 66.02 & 68.12 & 70.59 & +2.10 \\
Code & 0.754 & 51.26 & 52.10 & 55.47 & +0.84 \\
\bottomrule
\end{tabular}%
}
\end{table}

\clearpage
\section*{NeurIPS Paper Checklist}

\begin{enumerate}

\item {\bf Claims}
    \item[] Question: Do the main claims made in the abstract and introduction accurately reflect the paper's contributions and scope?
    \item[] Answer: \answerYes{}
    \item[] Justification: The abstract and introduction state the evaluated settings (verifiable rewards on 13 benchmarks; LLM-judge, human-preference, and continuous rewards; Qwen2.5 and Llama backbones from 1.5B to 8B), which match Sections~\ref{sec:main_results}--\ref{sec:generalization} and the appendices.
    \item[] Guidelines:
    \begin{itemize}
        \item The answer \answerNA{} means that the abstract and introduction do not include the claims made in the paper.
        \item The abstract and/or introduction should clearly state the claims made, including the contributions made in the paper and important assumptions and limitations. A \answerNo{} or \answerNA{} answer to this question will not be perceived well by the reviewers. 
        \item The claims made should match theoretical and experimental results, and reflect how much the results can be expected to generalize to other settings. 
        \item It is fine to include aspirational goals as motivation as long as it is clear that these goals are not attained by the paper. 
    \end{itemize}

\item {\bf Limitations}
    \item[] Question: Does the paper discuss the limitations of the work performed by the authors?
    \item[] Answer: \answerYes{}
    \item[] Justification: Appendix~\ref{appendix:limitations} discusses the scope of the evaluation, the frozen GNN, and the fixed sentence encoder.
    \item[] Guidelines:
    \begin{itemize}
        \item The answer \answerNA{} means that the paper has no limitation while the answer \answerNo{} means that the paper has limitations, but those are not discussed in the paper. 
        \item The authors are encouraged to create a separate ``Limitations'' section in their paper.
        \item The paper should point out any strong assumptions and how robust the results are to violations of these assumptions (e.g., independence assumptions, noiseless settings, model well-specification, asymptotic approximations only holding locally). The authors should reflect on how these assumptions might be violated in practice and what the implications would be.
        \item The authors should reflect on the scope of the claims made, e.g., if the approach was only tested on a few datasets or with a few runs. In general, empirical results often depend on implicit assumptions, which should be articulated.
        \item The authors should reflect on the factors that influence the performance of the approach. For example, a facial recognition algorithm may perform poorly when image resolution is low or images are taken in low lighting. Or a speech-to-text system might not be used reliably to provide closed captions for online lectures because it fails to handle technical jargon.
        \item The authors should discuss the computational efficiency of the proposed algorithms and how they scale with dataset size.
        \item If applicable, the authors should discuss possible limitations of their approach to address problems of privacy and fairness.
        \item While the authors might fear that complete honesty about limitations might be used by reviewers as grounds for rejection, a worse outcome might be that reviewers discover limitations that aren't acknowledged in the paper. The authors should use their best judgment and recognize that individual actions in favor of transparency play an important role in developing norms that preserve the integrity of the community. Reviewers will be specifically instructed to not penalize honesty concerning limitations.
    \end{itemize}

\item {\bf Theory assumptions and proofs}
    \item[] Question: For each theoretical result, does the paper provide the full set of assumptions and a complete (and correct) proof?
    \item[] Answer: \answerNA{}
    \item[] Justification: This is an empirical paper; no theoretical results or proofs are presented.
    \item[] Guidelines:
    \begin{itemize}
        \item The answer \answerNA{} means that the paper does not include theoretical results. 
        \item All the theorems, formulas, and proofs in the paper should be numbered and cross-referenced.
        \item All assumptions should be clearly stated or referenced in the statement of any theorems.
        \item The proofs can either appear in the main paper or the supplemental material, but if they appear in the supplemental material, the authors are encouraged to provide a short proof sketch to provide intuition. 
        \item Inversely, any informal proof provided in the core of the paper should be complemented by formal proofs provided in appendix or supplemental material.
        \item Theorems and Lemmas that the proof relies upon should be properly referenced. 
    \end{itemize}

    \item {\bf Experimental result reproducibility}
    \item[] Question: Does the paper fully disclose all the information needed to reproduce the main experimental results of the paper to the extent that it affects the main claims and/or conclusions of the paper (regardless of whether the code and data are provided or not)?
    \item[] Answer: \answerYes{}
    \item[] Justification: Appendix~\ref{appendix:hyperparams} provides complete hyperparameters for both GRPO training and GNN training. The GNN architecture is fully described in Section~\ref{sec:cross_domain_graph}, and all datasets are publicly available with statistics in Appendix~\ref{appendix:dataset_statistics}.
    \item[] Guidelines:
    \begin{itemize}
        \item The answer \answerNA{} means that the paper does not include experiments.
        \item If the paper includes experiments, a \answerNo{} answer to this question will not be perceived well by the reviewers: Making the paper reproducible is important, regardless of whether the code and data are provided or not.
        \item If the contribution is a dataset and\slash or model, the authors should describe the steps taken to make their results reproducible or verifiable. 
        \item Depending on the contribution, reproducibility can be accomplished in various ways. For example, if the contribution is a novel architecture, describing the architecture fully might suffice, or if the contribution is a specific model and empirical evaluation, it may be necessary to either make it possible for others to replicate the model with the same dataset, or provide access to the model. In general, releasing code and data is often one good way to accomplish this, but reproducibility can also be provided via detailed instructions for how to replicate the results, access to a hosted model (e.g., in the case of a large language model), releasing of a model checkpoint, or other means that are appropriate to the research performed.
        \item While NeurIPS does not require releasing code, the conference does require all submissions to provide some reasonable avenue for reproducibility, which may depend on the nature of the contribution. For example
        \begin{enumerate}
            \item If the contribution is primarily a new algorithm, the paper should make it clear how to reproduce that algorithm.
            \item If the contribution is primarily a new model architecture, the paper should describe the architecture clearly and fully.
            \item If the contribution is a new model (e.g., a large language model), then there should either be a way to access this model for reproducing the results or a way to reproduce the model (e.g., with an open-source dataset or instructions for how to construct the dataset).
            \item We recognize that reproducibility may be tricky in some cases, in which case authors are welcome to describe the particular way they provide for reproducibility. In the case of closed-source models, it may be that access to the model is limited in some way (e.g., to registered users), but it should be possible for other researchers to have some path to reproducing or verifying the results.
        \end{enumerate}
    \end{itemize}

\item {\bf Open access to data and code}
    \item[] Question: Does the paper provide open access to the data and code, with sufficient instructions to faithfully reproduce the main experimental results, as described in supplemental material?
    \item[] Answer: \answerYes{}
    \item[] Justification: Code, data splits, and trained checkpoints are released at the GitHub and Hugging Face links on the first page.
    \item[] Guidelines:
    \begin{itemize}
        \item The answer \answerNA{} means that paper does not include experiments requiring code.
        \item Please see the NeurIPS code and data submission guidelines (\url{https://neurips.cc/public/guides/CodeSubmissionPolicy}) for more details.
        \item While we encourage the release of code and data, we understand that this might not be possible, so \answerNo{} is an acceptable answer. Papers cannot be rejected simply for not including code, unless this is central to the contribution (e.g., for a new open-source benchmark).
        \item The instructions should contain the exact command and environment needed to run to reproduce the results. See the NeurIPS code and data submission guidelines (\url{https://neurips.cc/public/guides/CodeSubmissionPolicy}) for more details.
        \item The authors should provide instructions on data access and preparation, including how to access the raw data, preprocessed data, intermediate data, and generated data, etc.
        \item The authors should provide scripts to reproduce all experimental results for the new proposed method and baselines. If only a subset of experiments are reproducible, they should state which ones are omitted from the script and why.
        \item At submission time, to preserve anonymity, the authors should release anonymized versions (if applicable).
        \item Providing as much information as possible in supplemental material (appended to the paper) is recommended, but including URLs to data and code is permitted.
    \end{itemize}

\item {\bf Experimental setting/details}
    \item[] Question: Does the paper specify all the training and test details (e.g., data splits, hyperparameters, how they were chosen, type of optimizer) necessary to understand the results?
    \item[] Answer: \answerYes{}
    \item[] Justification: Section~4 describes the evaluation protocol; Appendix~\ref{appendix:hyperparams} provides all training hyperparameters (Tables~\ref{tab:hyperparams} and~\ref{tab:gnn_hyperparams}); Appendix~\ref{appendix:dataset_statistics} provides data splits for every dataset.
    \item[] Guidelines:
    \begin{itemize}
        \item The answer \answerNA{} means that the paper does not include experiments.
        \item The experimental setting should be presented in the core of the paper to a level of detail that is necessary to appreciate the results and make sense of them.
        \item The full details can be provided either with the code, in appendix, or as supplemental material.
    \end{itemize}

\item {\bf Experiment statistical significance}
    \item[] Question: Does the paper report error bars suitably and correctly defined or other appropriate information about the statistical significance of the experiments?
    \item[] Answer: \answerYes{}
    \item[] Justification: For R1-p, MemReward, and R1-Oracle, Tables~\ref{tab:main_results} and \ref{tab:ood_results} report the average score over six random seeds with its standard deviation. Table~\ref{tab:gt_ratio} reports the same statistics for the ground-truth-ratio conditions. Appendix~\ref{appendix:stats} reports two-tailed paired t-tests against R1-p for the conditions listed there.
    \item[] Guidelines:
    \begin{itemize}
        \item The answer \answerNA{} means that the paper does not include experiments.
        \item The authors should answer \answerYes{} if the results are accompanied by error bars, confidence intervals, or statistical significance tests, at least for the experiments that support the main claims of the paper.
        \item The factors of variability that the error bars are capturing should be clearly stated (for example, train/test split, initialization, random drawing of some parameter, or overall run with given experimental conditions).
        \item The method for calculating the error bars should be explained (closed form formula, call to a library function, bootstrap, etc.)
        \item The assumptions made should be given (e.g., Normally distributed errors).
        \item It should be clear whether the error bar is the standard deviation or the standard error of the mean.
        \item It is OK to report 1-sigma error bars, but one should state it. The authors should preferably report a 2-sigma error bar than state that they have a 96\% CI, if the hypothesis of Normality of errors is not verified.
        \item For asymmetric distributions, the authors should be careful not to show in tables or figures symmetric error bars that would yield results that are out of range (e.g., negative error rates).
        \item If error bars are reported in tables or plots, the authors should explain in the text how they were calculated and reference the corresponding figures or tables in the text.
    \end{itemize}

\item {\bf Experiments compute resources}
    \item[] Question: For each experiment, does the paper provide sufficient information on the computer resources (type of compute workers, memory, time of execution) needed to reproduce the experiments?
    \item[] Answer: \answerYes{}
    \item[] Justification: Appendix~\ref{appendix:hyperparams} specifies NVIDIA RTX PRO 6000 GPUs with BF16 precision. Appendix~\ref{appendix:cost} reports GNN training time (173s) and online inference overhead (0.03\%).
    \item[] Guidelines:
    \begin{itemize}
        \item The answer \answerNA{} means that the paper does not include experiments.
        \item The paper should indicate the type of compute workers CPU or GPU, internal cluster, or cloud provider, including relevant memory and storage.
        \item The paper should provide the amount of compute required for each of the individual experimental runs as well as estimate the total compute. 
        \item The paper should disclose whether the full research project required more compute than the experiments reported in the paper (e.g., preliminary or failed experiments that didn't make it into the paper). 
    \end{itemize}
    
\item {\bf Code of ethics}
    \item[] Question: Does the research conducted in the paper conform, in every respect, with the NeurIPS Code of Ethics \url{https://neurips.cc/public/EthicsGuidelines}?
    \item[] Answer: \answerYes{}
    \item[] Justification: The research conforms to the NeurIPS Code of Ethics. No human subjects are involved and no harmful applications are pursued. Appendix~\ref{appendix:limitations} discusses broader impacts.
    \item[] Guidelines:
    \begin{itemize}
        \item The answer \answerNA{} means that the authors have not reviewed the NeurIPS Code of Ethics.
        \item If the authors answer \answerNo, they should explain the special circumstances that require a deviation from the Code of Ethics.
        \item The authors should make sure to preserve anonymity (e.g., if there is a special consideration due to laws or regulations in their jurisdiction).
    \end{itemize}

\item {\bf Broader impacts}
    \item[] Question: Does the paper discuss both potential positive societal impacts and negative societal impacts of the work performed?
    \item[] Answer: \answerYes{}
    \item[] Justification: Appendix~\ref{appendix:limitations} discusses the positive impact of reducing annotation cost for RL-based LLM training and notes that downstream deployments should incorporate appropriate safeguards.
    \item[] Guidelines:
    \begin{itemize}
        \item The answer \answerNA{} means that there is no societal impact of the work performed.
        \item If the authors answer \answerNA{} or \answerNo, they should explain why their work has no societal impact or why the paper does not address societal impact.
        \item Examples of negative societal impacts include potential malicious or unintended uses (e.g., disinformation, generating fake profiles, surveillance), fairness considerations (e.g., deployment of technologies that could make decisions that unfairly impact specific groups), privacy considerations, and security considerations.
        \item The conference expects that many papers will be foundational research and not tied to particular applications, let alone deployments. However, if there is a direct path to any negative applications, the authors should point it out. For example, it is legitimate to point out that an improvement in the quality of generative models could be used to generate Deepfakes for disinformation. On the other hand, it is not needed to point out that a generic algorithm for optimizing neural networks could enable people to train models that generate Deepfakes faster.
        \item The authors should consider possible harms that could arise when the technology is being used as intended and functioning correctly, harms that could arise when the technology is being used as intended but gives incorrect results, and harms following from (intentional or unintentional) misuse of the technology.
        \item If there are negative societal impacts, the authors could also discuss possible mitigation strategies (e.g., gated release of models, providing defenses in addition to attacks, mechanisms for monitoring misuse, mechanisms to monitor how a system learns from feedback over time, improving the efficiency and accessibility of ML).
    \end{itemize}
    
\item {\bf Safeguards}
    \item[] Question: Does the paper describe safeguards that have been put in place for responsible release of data or models that have a high risk for misuse (e.g., pre-trained language models, image generators, or scraped datasets)?
    \item[] Answer: \answerNA{}
    \item[] Justification: The paper proposes a training framework rather than releasing a new pretrained model or scraped dataset. The fine-tuned models are standard Qwen2.5 variants on public benchmarks.
    \item[] Guidelines:
    \begin{itemize}
        \item The answer \answerNA{} means that the paper poses no such risks.
        \item Released models that have a high risk for misuse or dual-use should be released with necessary safeguards to allow for controlled use of the model, for example by requiring that users adhere to usage guidelines or restrictions to access the model or implementing safety filters. 
        \item Datasets that have been scraped from the Internet could pose safety risks. The authors should describe how they avoided releasing unsafe images.
        \item We recognize that providing effective safeguards is challenging, and many papers do not require this, but we encourage authors to take this into account and make a best faith effort.
    \end{itemize}

\item {\bf Licenses for existing assets}
    \item[] Question: Are the creators or original owners of assets (e.g., code, data, models), used in the paper, properly credited and are the license and terms of use explicitly mentioned and properly respected?
    \item[] Answer: \answerYes{}
    \item[] Justification: All datasets and models are properly cited. All benchmarks used are publicly available under permissive licenses. TruthfulQA, Chatbot Arena, and HotpotQA are cited at first use and used under their respective licenses and terms of use, and the Llama-3 models are used under the Llama 3.1 and 3.2 Community Licenses.
    \item[] Guidelines:
    \begin{itemize}
        \item The answer \answerNA{} means that the paper does not use existing assets.
        \item The authors should cite the original paper that produced the code package or dataset.
        \item The authors should state which version of the asset is used and, if possible, include a URL.
        \item The name of the license (e.g., CC-BY 4.0) should be included for each asset.
        \item For scraped data from a particular source (e.g., website), the copyright and terms of service of that source should be provided.
        \item If assets are released, the license, copyright information, and terms of use in the package should be provided. For popular datasets, \url{https://paperswithcode.com/datasets} has curated licenses for some datasets. Their licensing guide can help determine the license of a dataset.
        \item For existing datasets that are re-packaged, both the original license and the license of the derived asset (if it has changed) should be provided.
        \item If this information is not available online, the authors are encouraged to reach out to the asset's creators.
    \end{itemize}

\item {\bf New assets}
    \item[] Question: Are new assets introduced in the paper well documented and is the documentation provided alongside the assets?
    \item[] Answer: \answerYes{}
    \item[] Justification: The released code and data are documented in the GitHub and Hugging Face repositories linked on the first page.
    \item[] Guidelines:
    \begin{itemize}
        \item The answer \answerNA{} means that the paper does not release new assets.
        \item Researchers should communicate the details of the dataset\slash code\slash model as part of their submissions via structured templates. This includes details about training, license, limitations, etc. 
        \item The paper should discuss whether and how consent was obtained from people whose asset is used.
        \item At submission time, remember to anonymize your assets (if applicable). You can either create an anonymized URL or include an anonymized zip file.
    \end{itemize}

\item {\bf Crowdsourcing and research with human subjects}
    \item[] Question: For crowdsourcing experiments and research with human subjects, does the paper include the full text of instructions given to participants and screenshots, if applicable, as well as details about compensation (if any)? 
    \item[] Answer: \answerNA{}
    \item[] Justification: No crowdsourcing or human subjects research is involved.
    \item[] Guidelines:
    \begin{itemize}
        \item The answer \answerNA{} means that the paper does not involve crowdsourcing nor research with human subjects.
        \item Including this information in the supplemental material is fine, but if the main contribution of the paper involves human subjects, then as much detail as possible should be included in the main paper. 
        \item According to the NeurIPS Code of Ethics, workers involved in data collection, curation, or other labor should be paid at least the minimum wage in the country of the data collector. 
    \end{itemize}

\item {\bf Institutional review board (IRB) approvals or equivalent for research with human subjects}
    \item[] Question: Does the paper describe potential risks incurred by study participants, whether such risks were disclosed to the subjects, and whether Institutional Review Board (IRB) approvals (or an equivalent approval/review based on the requirements of your country or institution) were obtained?
    \item[] Answer: \answerNA{}
    \item[] Justification: No human subjects research is involved.
    \item[] Guidelines:
    \begin{itemize}
        \item The answer \answerNA{} means that the paper does not involve crowdsourcing nor research with human subjects.
        \item Depending on the country in which research is conducted, IRB approval (or equivalent) may be required for any human subjects research. If you obtained IRB approval, you should clearly state this in the paper. 
        \item We recognize that the procedures for this may vary significantly between institutions and locations, and we expect authors to adhere to the NeurIPS Code of Ethics and the guidelines for their institution. 
        \item For initial submissions, do not include any information that would break anonymity (if applicable), such as the institution conducting the review.
    \end{itemize}

\item {\bf Declaration of LLM usage}
    \item[] Question: Does the paper describe the usage of LLMs if it is an important, original, or non-standard component of the core methods in this research? Note that if the LLM is used only for writing, editing, or formatting purposes and does \emph{not} impact the core methodology, scientific rigor, or originality of the research, declaration is not required.
    \item[] Answer: \answerYes{}
    \item[] Justification: LLMs are the policies trained by MemReward (Section~4.1), and the TruthfulQA experiment uses DeepSeek-V4-Flash as a judge (Appendix~\ref{appendix:reward_sources}). LLMs were otherwise used only for editing the writing.
    \item[] Guidelines:
    \begin{itemize}
        \item The answer \answerNA{} means that the core method development in this research does not involve LLMs as any important, original, or non-standard components.
        \item Please refer to our LLM policy in the NeurIPS handbook for what should or should not be described.
    \end{itemize}

\end{enumerate}

\end{document}